\documentclass[journal]{IEEEtran}

\usepackage{cite}
\ifCLASSINFOpdf 

  \usepackage[pdftex]{graphicx}
\else
  \usepackage[dvips]{graphicx}
\fi
\usepackage{amsmath,amssymb,amsfonts}
\usepackage{array}

\usepackage{fancyhdr}
\usepackage{booktabs} 
\usepackage{multirow,multicol}
\usepackage{color,xcolor}
\usepackage{graphicx}
\usepackage{epstopdf}
\usepackage{booktabs}
\usepackage{cite}
\usepackage[Symbol]{upgreek}
\usepackage{psfrag}
\usepackage{setspace}
\usepackage{acronym}
\usepackage{arydshln}

\usepackage{bm}
\usepackage{verbatim}

\usepackage{algorithm}
\usepackage{algorithmicx}
\usepackage{algpseudocode}

\ifCLASSOPTIONcompsoc
  \usepackage[caption=false,font=normalsize,labelfont=sf,textfont=sf]{subfig}
\else
  \usepackage[caption=false,font=footnotesize]{subfig}
\fi
\usepackage{fixltx2e}

\usepackage{stfloats}

\ifCLASSOPTIONcaptionsoff
  \usepackage[nomarkers]{endfloat}
 \let\MYoriglatexcaption\caption
 \renewcommand{\caption}[2][\relax]{\MYoriglatexcaption[#2]{#2}}
\fi

\usepackage{url}

\DeclareMathAlphabet{\mathsfbr}{OT1}{cmss}{m}{n}
\SetMathAlphabet{\mathsfbr}{bold}{OT1}{cmss}{bx}{n}
\DeclareRobustCommand{\msf}[1]{%
  \ifcat\noexpand#1\relax\msfgreek{#1}\else\mathsfbr{#1}\fi
}

\makeatletter
\newcommand{\msfgreek}[1]{\csname s\expandafter\@gobble\string#1\endcsname}
\makeatother

\DeclareFontEncoding{LGR}{}{} 
\DeclareSymbolFont{sfgreek}{LGR}{cmss}{m}{n}
\SetSymbolFont{sfgreek}{bold}{LGR}{cmss}{bx}{n}
\DeclareMathSymbol{\salpha}{\mathord}{sfgreek}{`a}
\DeclareMathSymbol{\sbeta}{\mathord}{sfgreek}{`b}
\DeclareMathSymbol{\sgamma}{\mathord}{sfgreek}{`g}
\DeclareMathSymbol{\sdelta}{\mathord}{sfgreek}{`d}
\DeclareMathSymbol{\sepsilon}{\mathord}{sfgreek}{`e}
\DeclareMathSymbol{\szeta}{\mathord}{sfgreek}{`z}
\DeclareMathSymbol{\seta}{\mathord}{sfgreek}{`h}
\DeclareMathSymbol{\stheta}{\mathord}{sfgreek}{`j}
\DeclareMathSymbol{\siota}{\mathord}{sfgreek}{`i}
\DeclareMathSymbol{\skappa}{\mathord}{sfgreek}{`k}
\DeclareMathSymbol{\slambda}{\mathord}{sfgreek}{`l}
\DeclareMathSymbol{\smu}{\mathord}{sfgreek}{`m}
\DeclareMathSymbol{\snu}{\mathord}{sfgreek}{`n}
\DeclareMathSymbol{\sxi}{\mathord}{sfgreek}{`x}
\DeclareMathSymbol{\somicron}{\mathord}{sfgreek}{`o}
\DeclareMathSymbol{\spi}{\mathord}{sfgreek}{`p}
\DeclareMathSymbol{\srho}{\mathord}{sfgreek}{`r}
\DeclareMathSymbol{\ssigma}{\mathord}{sfgreek}{`s}
\DeclareMathSymbol{\stau}{\mathord}{sfgreek}{`t}
\DeclareMathSymbol{\supsilon}{\mathord}{sfgreek}{`u}
\DeclareMathSymbol{\sphi}{\mathord}{sfgreek}{`f}
\DeclareMathSymbol{\schi}{\mathord}{sfgreek}{`q}
\DeclareMathSymbol{\spsi}{\mathord}{sfgreek}{`y}
\DeclareMathSymbol{\somega}{\mathord}{sfgreek}{`w}

\DeclareMathSymbol{\svarsigma}{\mathord}{sfgreek}{`c}

\DeclareMathSymbol{\sGamma}{\mathalpha}{sfgreek}{`G}
\DeclareMathSymbol{\sDelta}{\mathalpha}{sfgreek}{`D}
\DeclareMathSymbol{\sTheta}{\mathalpha}{sfgreek}{`J}
\DeclareMathSymbol{\sLambda}{\mathalpha}{sfgreek}{`L}
\DeclareMathSymbol{\sXi}{\mathalpha}{sfgreek}{`X}
\DeclareMathSymbol{\sPi}{\mathalpha}{sfgreek}{`P}
\DeclareMathSymbol{\sSigma}{\mathalpha}{sfgreek}{`S}
\DeclareMathSymbol{\sUpsilon}{\mathalpha}{sfgreek}{`U}
\DeclareMathSymbol{\sPhi}{\mathalpha}{sfgreek}{`F}
\DeclareMathSymbol{\sPsi}{\mathalpha}{sfgreek}{`Y}
\DeclareMathSymbol{\sOmega}{\mathalpha}{sfgreek}{`W}

\DeclareRobustCommand{\mcal}[1]{%
  \ifcat\noexpand#1\relax\mathnormal{#1}\else\cal{#1}\fi
}
\DeclareRobustCommand{\BM}[1]{%
  \ifcat\noexpand#1\relax\bm{\boldUppercaseItalicGreek{#1}}\else\bm{#1}\fi
}
\makeatletter
\newcommand{\boldUppercaseItalicGreek}[1]{\csname var\expandafter\@gobble\string#1\endcsname}
\makeatother
\newcommand{\rv}[1]{\MakeLowercase{\msf{#1}}}
\newcommand{\RV}[1]{\bm{\MakeLowercase{\msf{#1}}}}

\newcommand{\V}[1]{\bm{#1}}

\newtheorem{proof}{Proof}

\begin{document}
%
\title{A Deep Generative Model for Synthesizing Labeled Wireless Signals}
%
%
%

\author{Yuxiao~Li,
        Keke~Hu,
        Santiago~Mazuelas,
        and~Yuan~Shen
\thanks{Yuxiao Li and Santiago Mazuelas are with the Basque Center for Applied Mathematics (BCAM), 48009 Bilbao, Spain, and also with the IKERBASQUE Foundation for Science, 48009 Bilbao, Spain (e-mails: yli@bcamath.org, smazuelas@bcamath.org).}
\thanks{Keke Hu is with the College of Semiconductors (College of Integrated Circuits), Changsha 410082, China. (e-mail: hukeke@hnu.edu.cn)}
\thanks{Yuan Shen is with the Department of Electronic Engineering, Tsinghua University, Beijing 100190, China, and also with the Beijing National Research Center for Information Science and Technology, Beijing 100084, China. (e-mail: shenyuan\_ee\@tsinghua.edu.cn).}
}

\maketitle

\begin{abstract}

Wireless signals with position-related labels are pivotal for both performance evaluation and model training in the realm of wireless sensing.
However, acquiring real-world datasets is often challenged by significant measurement and labeling costs. Traditional methods for synthesizing labeled wireless signals typically rely on environmental models, leading to extensive hyper-parameter tuning and inadequate realism for comprehensive model training purposes.
To address these limitations, we introduce a novel deep learning (DL)-based method, namely Inter-Instance Generative Adversarial Networks (IIns-GAN), to generate realistic labeled wireless signals.
The generated signals are particularly adaptive to different environment scenarios and well-suited for various model training tasks, including distance estimation and environment identification.
We have conducted extensive experiments on public Ultra-Wideband (UWB) datasets to evaluate the realism and utility of the generated signals.
The results demonstrate that the signals generated by IIns-GAN mirror the physical characteristics of real-world measurements, and significantly contribute to the improvement of model training in diverse wireless sensing tasks.

\end{abstract}


%
\IEEEpeerreviewmaketitle

\section{Introduction}
\label{sec:intro}

The availability of labeled wireless signals is crucial for research and performance evaluation across a wide range of applications, including localization \cite{obeidat2021review,BurRavRao:J20}, Internet of Things (IoT) \cite{LiDaZha:J15,ZanBuiCas:J14}, and wearable technology \cite{ha2021wistress,zhang2023embracing}. With the proliferation of machine learning in wireless systems \cite{wang2017deep}, large datasets with labeled wireless signals are required to train high-performing data-driven models \cite{zappone2019wireless,MazConAllWin:J18,huang2022indoor}.
Their importance is further amplified with the emergence of deep learning (DL) and artificial intelligence techniques \cite{KanRap:J21,Kanhere2021PositionLF,himeur2022ai}. In particular, a tremendous amount of data-driven methods are being developed in wireless sensing tasks such as localization estimation and environment sensing \cite{azhari2023deep,li2020wi,chen2021deep}. These methods rely on a well-built dataset to learn the complicated relationships between received high-dimensional signals and position-related labels.
However, acquiring suitable wireless signals with accurate annotations remain challenging. Signals must cover various environments \cite{molisch2012wireless} and distance ranges with balanced representation, requiring extensive measurement campaigns.

\begin{figure}[htbp]
\centering
\includegraphics[height=1.6in]{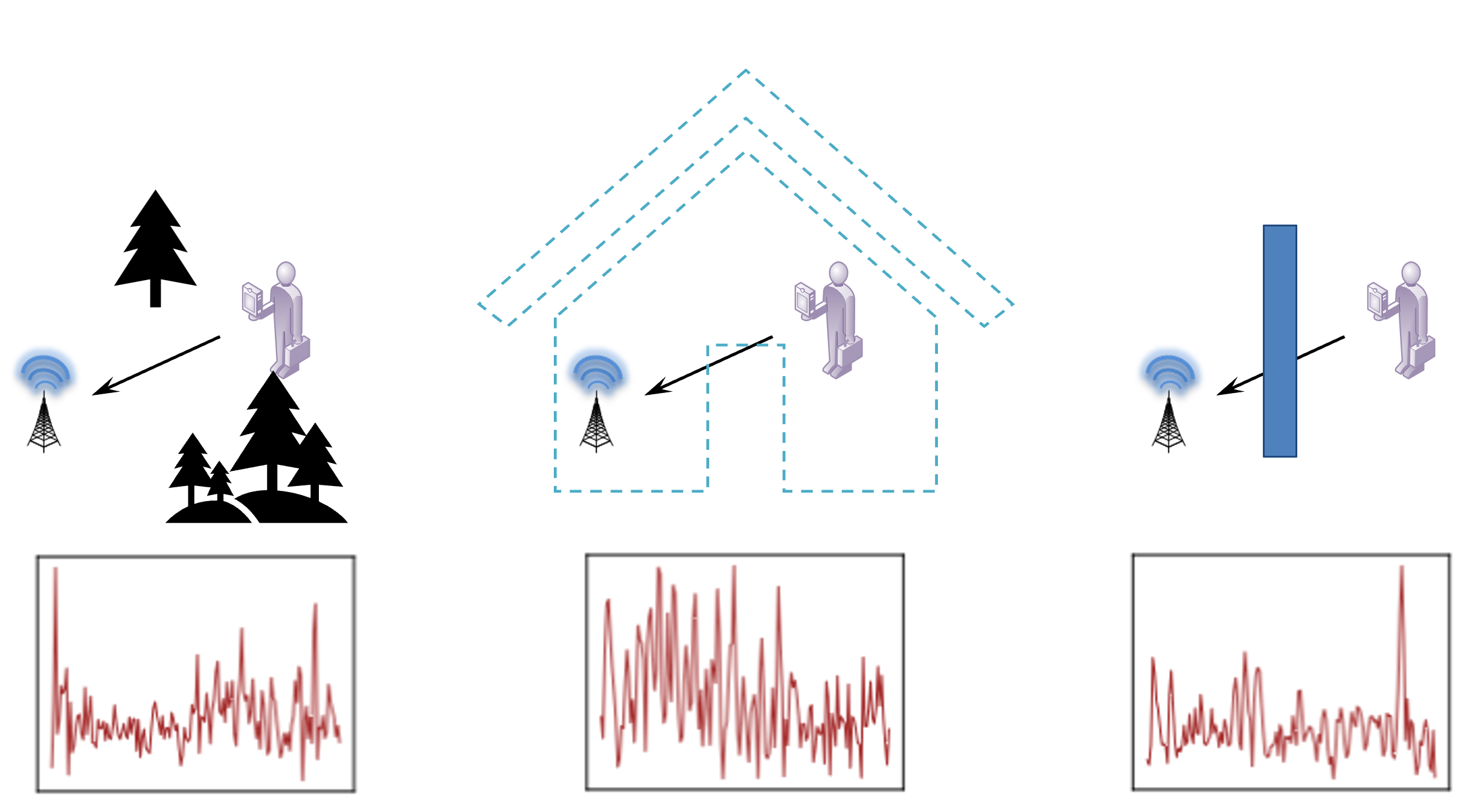}%
\caption{Signal propagation in different environments: outdoor (LOS), indoor (multipath), and NLOS with wood and metal obstacles. Corresponding received signal waveforms illustrate the impact on signal quality.}
\label{fig:intro}
\end{figure}

The acquisition remains challenging of wireless signal datasets with diverse, balanced instances across different ranges and environments.
Specifically, performing extensive measurement campaigns is time-consuming and labor-intensive \cite{zhou2022automatic,khodjaev2010survey,LiMazShe:C22_3,LiMazShe:C22_4}. Obtaining accurate labels for training data-driven models requires cumbersome human labor \cite{shahmansoori20155g,li2021deep}.
Moreover, acquiring positional labels in harsh environments like indoor areas is even more challenging and sometimes infeasible. For example, the ground-truth distance between a transmitter and a receiver in a dense office environment is hard to measure precisely \cite{obeidat2021review,LiMazShe:C22}. These challenges limit the dataset construction from real-world measurements for wireless sensing tasks.


Traditional methods of signal synthesis, such as Rayleigh fading models and ray-tracing techniques, often rely on physical or statistic models \cite{rappaport2002wireless}. Such models make simplified assumptions about the environment, leading to signals that lack the complexity and realism of actual wireless scenarios. Recent works improve the model design for signal synthesis in more diverse environments \cite{Ihler2005NonparametricBP,Deng2014MmwaveMC,zajic2009three}. However, these methods requires delicate fine-tuning of hyper-parameters in different scenarios, impractical in large dataset generation. Such gap in realism presents significant challenges, particularly in the context of training data-driven models and introducing the latest DL techniques to evolving wireless applications.

Deep Generative Models (DGMs), such as variational auto-encoders (VAEs) \cite{KinWel:C13} and generative adversarial networks (GANs) \cite{Goodfellow2014GenerativeAN}, have achieved an impressive success in producing highly realistic outputs. Instead of defining a pre-fixed model, DGMs directly learn data distributions from real-world datasets via deep neural networks. Equipped with modern GPUs, DGMs are capable to generate realistic data samples with high fidelity \cite{haneda2021radio,zakaria2021developed,aldossari2019machine}. The remarkable advancements of DGMs have been shown in computer vision (CV) \cite{Isola2017ImagetoImageTW,li2022variational} and natural language processing (NLP) \cite{fu2018style,chou2019one}. Therefore, such methods offer a promising solution to realistic wireless signal generation.

In this paper, we propose a DGM-based framework for wireless signal generation with position-related labels, as depicted in Fig. \ref{fig:intro}. The framework is developed using a latent variable model (LVM) with variational inference (VI) techniques, and implemented by generative adversarial networks (GANs).
The main contributions in the paper are as follows:
\begin{itemize}
    \item We propose a VI method to conduct signal generation on a LVM. The model accounts for realistic signal generation conditioned on different position-related features.
    \item We design IIns-GAN, a DGM-based neural network to implement labeled signal generation. The network is capable of generating realistic signals with various distance and environment labels.
    \item We validate the realism of the generated labeled signals and their utility in enhancing modeling training for wireless sensing applications.
\end{itemize}

The rest of the paper is organized as follows. Section \ref{sec:problem} introduces the problem of labeled signal generation and the proposed LVM. Section \ref{sec:model} develops the variational inference method to conduct signal estimation conditioned on position-related features. Section \ref{sec:network} proposes the IIns-GAN framework to implement the inference for labeled signal generation, encompassing both label-based synthesis and signal-based translation. Section \ref{sec:discussion} provides a brief discussion on additional downstream tasks, connections and implications of the proposed IIns-GAN. Experimental results and analyses are presented in Section \ref{sec:exp}. Finally, Section VI concludes the paper.

\textit{Notations:} random variables (RVs) are displayed in sans serif, upright fonts and their realizations in serif, italic fonts; vectors are denoted by bold lowercase letters; a RV and its realization are denoted by $\mathrm{x}$ and $x$; a random vector and its realization are denoted by $\mathbf{\mathrm{x}}$ and $\mathbf{x}$; $\mathbf{x}[j]$ denotes the $j$th component of the vector $\mathbf{x}$; $f_{\mathrm{x}}(x)$ and, for brevity when possible, $f(\mathbf{x})$ denote the Radon-Nikodym derivative of a RV $\mathbf{\mathrm{x}}$ with respect to the base measure, e.g., $f(\mathbf{x})$ denotes a probability density function (PDF) in case of a continuous RV $\mathbf{\mathrm{x}}$; $f(\mathbf{x}|\mathbf{z})$ denotes either the conditional distribution of $\mathbf{\mathrm{x}}$ given $\mathbf{\mathrm{z}}=\mathbf{z}$ for a RV $\mathbf{\mathrm{z}}$ or the distribution of $\mathbf{\mathrm{x}}$ parameterized by $\mathbf{\mathrm{z}}$ for a parameter $\mathbf{z}$; $\varphi(\mathbf{x};\boldsymbol{\mu}, \boldsymbol{\Sigma})$ denotes the PDF of a Gaussian RV $\mathbf{\mathrm{x}}$ with mean $\boldsymbol{\mu}$ and covariance matrix $\boldsymbol{\Sigma}$; $\mathbb{E}\{\cdot\}$, $\mathbb{V}\{\cdot\}$, and $\mathbb{P}\{\cdot\}$ denote, respectively, the expectation, variance, and propability of the argument, and $\mathbb{E}_{\mathrm{x}}\{\cdot\}$ denotes the expectation with respect to RV $\mathrm{x}$; $[\cdot]^T$ denotes the transpose of the argument.
Sets are denoted by calligraphic fonts, e.g., $\mathcal{Y}$. The norm of a vector $\mathbf{u}$ is denoted by $\Vert \mathbf{u}\Vert$. 

\section{Proposed Framework for Signal Generation}
\label{sec:problem}

In this section, we present the problem formulation and the hierarchical LVM for signal generation. The introduced LVM describes the relationships among different variables and enables to generate realistic signals with selected distance and environment features.

\subsection{Problem Formulation}
\label{sec:pro_model}

\begin{figure}[!t]
\centering
{\includegraphics[height=1.6in]{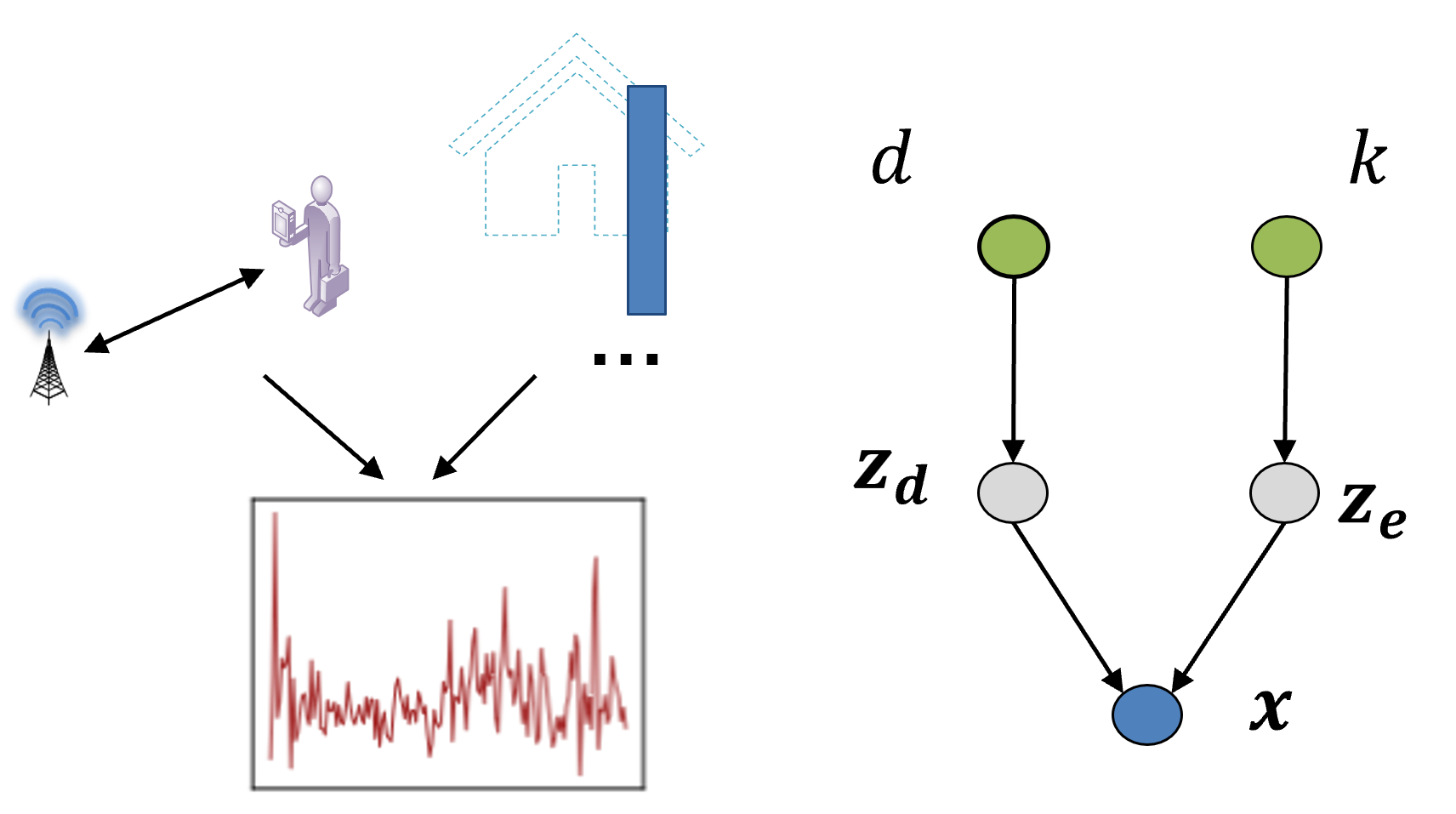}}%
\caption{Graphical model for the signal measurement $\V{x}$, distance and environmental features $\V{z}_{\text{d}}$, $\V{z}_{\text{e}}$, as well as distance and environmental labels $d$ and $k$.} 
\label{fig:graphics}
\end{figure}


Consider a wireless signal $\V{x}\in\mathbb{R}^M$ measured between a transmitter and a receiver, where each component of the vector $\V{x}$ is the value of sampled amplitude. The environment label $k\in\mathcal{K}=\{0, 1, \ldots, K-1\}$ characterizes the type of surrounding propagation environment. For example, $\mathcal{K}$ can be constructed to indicate LOS and NLOS conditions, i.e., ${k}\in\mathcal{K}_{\text{NLOS}}=\{0, 1\}$ with $k=0$ for LOS and $k=1$ for NLOS conditions. The distance label $d\in[0, d_{\text{max}}]$ represents the real distance between the transmitter and the receiver.
In this paper we consider the distance as the positional metric of interest, but the framework can be extended to other positional metrics like angle, velocity, etc.

Wireless signal generation consists of obtaining a realistic signal $\V{x}_{d,k}\in \mathbb{R}^M$ that reflects features associated to distance label $d$ and environment label $k$.

\subsection{Latent Variable Model}
\label{sec:pro_lvm}


We utilize the hierarchical LVM proposed in \cite{LiMazShe:J23} to characterize wireless signals in complex environments. A wireless signal $\V{x}$ inherently encodes positional and environmental information represented by the distance and environment labels, as dipicted in Fig.~\ref{fig:graphics}. The LVM contains multiple levels of latent variables that disentangle positional and environmental factors of variation in the signal $\V{x}$. Specifically, the LVM utilizes latent variables $\RV{z}_{\text{d}}$ and $\RV{z}_{\text{e}}$ to model lower-level distance and environment signal features, respectively. The distance label $\rv{d}$ and environment label $\rv{k}$ are higher-level latent variables that generate the feature variables through distributions $p(\V{z}_{\text{d}}|d)$ and $p(\V{z}_{\text{e}}|k)$. The structural relationships of variables of the LVM are given by the following modeling assumptions.

\begin{enumerate}
    \item In the absence of measurements, the distance $d$ and environment scenario $k$ are independent,
\begin{equation}  \label{eq:ass_1-prior}
    p(d,k) = p(d)p(k)
\end{equation}
    \item The latent variable $\V{z}_{\text{d}}$ is independent of $k$ and $\V{z}_{\text{e}}$ given the distance $d$, and the latent variable $\V{z}_{\text{e}}$ is independent of $d$ and $\V{z}_{\text{d}}$ given the environment label $k$,
\begin{equation}  \label{eq:ass_1-latent}
    p(\V{z}_{\text{d}},\V{z}_{\text{e}}|d, k) = p(\V{z}_{\text{d}}|d)p(\V{z}_{\text{e}}|k)
\end{equation}
    \item The signal measurement $\V{x}$ is independent of $d$ and $k$ given $\V{z}_{\text{d}}$ and $\V{z}_{\text{e}}$,
\begin{equation}  \label{eq:ass_1-obs}
    p(\V{x}|\V{z}_{\text{d}},\V{z}_{\text{e}},d,k) = p(\V{x}|\V{z}_{\text{d}},\V{z}_{\text{e}})
\end{equation}
\end{enumerate}

These assumptions provide disentanglement between positional and environmental sources of variation in complex wireless signals. The inference and generative modeling enabled by this LVM make it suitable for controllable signal generation given specific labels $d$ and $k$.



Given the LVM, the generation of wireless signal $\V{x}_{d,k}$ can be achieved via two types of techniques: 1) label-based synthesis and 2) signal-based translation.
On one hand, signal synthesis can be conducted by estimating the signal $\V{x}_{d,k}$ directly from specific $d$ and $k$ labels, following the LVM generation flow from labels to features and the observed signals.
On the other hand, signal-based translation can be conducted by first estimating features from a reference signal $\V{x}'$ via the LVM inference flow, and then recompose the desired features to generate the new signal $\V{x}_{d,k}$ via the generation flow.


\section{Variational Inference Method for Signal Generation}
\label{sec:model}

In this section, we propose the VI method for signal generation. We first outline the basic VI framework on the LVM. We then introduce the implicit distribution assumption (that sets our approach apart) to specify the VI objectives. Finally, we derive the corresponding empirical objective functions based on the assumption for VI approximation.

\subsection{Variational Inference Method Framework}
\label{sec:m_vi}

According to the LVM above, signal generation requires the estimation of latent variables $\V{z}_{\text{d}}$ and $\V{z}_{\text{e}}$, either conditioned on given signal $\V{x}$ or labels $d$ and ${k}$. Based on the VI method, we construct the variational distribution $q(\V{z}_{\text{d}},\V{z}_{\text{e}}|\V{x};\boldsymbol{\phi})$ to approximate the intractable posterior $p(\V{z}_{\text{d}},\V{z}_{\text{e}}|\V{x})$, and variational distributions $q(\V{z}_{\text{d}}|{d};\boldsymbol{\varphi}_{\text{d}}^{'})$ and $q(\V{z}_{\text{e}}|{k};\boldsymbol{\varphi}_{\text{e}}^{'})$ to approximate the intractable posteriors $p(\V{z}_{\text{d}}|d)$ and $p(\V{z}_{\text{e}}|k)$, respectively. Specifically, we take the variational distributions from parametric families satisfying the LVM with their PDFs differentiable almost everywhere with respect to variables and parameters. The specific parametric distribution assumptions are given in the following.

\textit{Assumption 1.} (Parametric Variational Distributions) In the following we use the assumptions:
\begin{enumerate}
    \item According to the mean-field assumption in VI \cite{BleKucMcA:J17}, we assume that the variational distributions of the two latent variables are independent as follows
\begin{equation}  \label{eq:ass_3}
    q(\V{z}_{\text{d}},\V{z}_{\text{e}}|\V{x};\boldsymbol{\phi}) = q(\V{z}_{\text{d}}|\V{x};\boldsymbol{\phi}_{\text{d}})q(\V{z}_{\text{e}}|\V{x};\boldsymbol{\phi}_{\text{e}})
\end{equation}
\noindent with $\boldsymbol{\phi}=\{\boldsymbol{\phi}_{\text{d}},\boldsymbol{\phi}_{\text{e}}\}$ for convenience.
    \item For network parameter learning, we assume that the conditional distribution on measurements variable $p(\V{x}|\V{z}_{\text{d}},\V{z}_{\text{e}};\boldsymbol{\theta})$ is from a parametric family of distributions with parameters $\boldsymbol{\theta}$. The conditional distributions on label variables $p(d|\V{z}_{\text{d}};\boldsymbol{\varphi}_{\text{d}})$ and $p(k|\V{z}_{\text{e}};\boldsymbol{\varphi}_{\text{e}})$ are from parametric families with $\boldsymbol{\varphi}_{\text{d}}$ and $\boldsymbol{\varphi}_{\text{e}}$. We denote $\boldsymbol{\varphi}=\{\boldsymbol{\varphi}_{\text{d}},\boldsymbol{\varphi}_{\text{e}}\}$ for convenience.
    \item We further construct distributions $q(\V{z}_{\text{d}}|d;\boldsymbol{\varphi}^{'}_{\text{d}})$, $q(\V{z}_{\text{k}}|k;\boldsymbol{\varphi}^{'}_{\text{k}})$ with parameters $\boldsymbol{\varphi}^{'}_{\text{d}}$, $\boldsymbol{\varphi}^{'}_{\text{k}}$ to approximate $p(\V{z}_{\text{d}}|d)$ and $p(\V{z}_{\text{k}}|k)$. We denote $\boldsymbol{\varphi}^{'}=\{\boldsymbol{\varphi}^{'}_{\text{d}},\boldsymbol{\varphi}^{'}_{\text{k}}\}$ for convenience.
    Likewise, their PDFs are assumed to be differentiable almost everywhere with respect to both conditioned variables and parameters.
\end{enumerate}

According to \textit{Proposition 1} in \cite{LiMazShe:J23},  the estimation of distribution parameters $\boldsymbol{\phi}$, $\boldsymbol{\theta}$, and $\boldsymbol{\varphi}$ can be conducted by via maximizing the following evidence lower bound (ELBO), where for each instance-labels pair $(\V{x},d ,k)$,
\begin{equation}  \label{eq:bound_pro}
    \begin{aligned}
    \mathcal{L}_{\text{ELBO}}(\V{x},d,k;\boldsymbol{\phi},\boldsymbol{\theta},\boldsymbol{\varphi}) & = \mathbb{E}_{q(\V{z}_{\text{d}},\V{z}_{\text{e}}|\V{x};\boldsymbol{\phi})}\big\{\log p(\V{x}|\V{z}_{\text{d}},\V{z}_{\text{e}};\boldsymbol{\theta})\big\} \\
    &\quad - \operatorname{D}_{\text{KL}}\big(q(\V{z}_{\text{d}},\V{z}_{\text{e}}|\V{x};\boldsymbol{\phi})\big|\big|p(\V{z}_{\text{d}},\V{z}_{\text{e}})\big)  \\
    &\quad + \mathbb{E}_{q(\V{z}_{\text{d}}|\V{x};\boldsymbol{\phi})}\big\{\log p(d|\V{z}_{\text{d}};\boldsymbol{\varphi})\big\}  \\
    &\quad + \mathbb{E}_{q(\V{z}_{\text{e}}|\V{x};\boldsymbol{\phi})}\big\{\log p(k|\V{z}_{\text{e}};\boldsymbol{\varphi})\big\}  \\
    &\leq \log~p(\V{x}, d, k)
    \end{aligned}
\end{equation}
In addition, the bound in \eqref{eq:bound_pro} holds with equality if and only if $q(\V{z}_{\text{d}},\V{z}_{\text{e}}|\V{x},d,k)$ matches the true posterior $p(\V{z}_{\text{d}},\V{z}_{\text{e}}|\V{x},d,k)$ perfectly, i.e., $q(\V{z}_{\text{d}},\V{z}_{\text{e}}|\V{x},d,k)=p(\V{z}_{\text{d}},\V{z}_{\text{e}}|\V{x},d,k)$ for the instance-labels pair $(\V{x}, d, k)$.

The estimation of distribution parameters $\boldsymbol{\varphi}^{'}$ can then be conducted by via minimizing the KL divergence between approximated and true distributions. The following proposition gives the objective function that enables such estimation.

\textit{Proposition 1:} (Feature Generation Objective) If \textit{Assumptions 1-2} are satisfied, we have the following equation for each instance-labels pair $(\V{x},d ,k)$,
\begin{equation} \label{eq:kl_gen}
    \begin{aligned}
        &\quad \mathbb{L}_{\text{KL}}(\V{x},d,k; \boldsymbol{\varphi}^{'}) \\
        &=\operatorname{D}_{\text{KL}}\big(q(\V{z}_{\text{d}},\V{z}_{\text{e}}|d,k;\boldsymbol{\varphi}^{'})\big|\big|p(\V{z}_{\text{d}},\V{z}_{\text{e}}|d,k)\big)  \\
        &= \operatorname{D}_{\text{KL}}\big(q(\V{z}_{\text{d}}|d;\boldsymbol{\varphi}_{\text{d}}^{'})\big|\big|p(\V{z}_{\text{d}}|d)\big) + \operatorname{D}_{\text{KL}}\big(q(\V{z}_{\text{e}}|k;\boldsymbol{\varphi}_{\text{e}}^{'})\big|\big|p(\V{z}_{\text{e}}|k)\big)  \\
        &= \operatorname{D}_{\text{KL}}\big(q(\V{z}_{\text{d}}|d;\boldsymbol{\varphi}_{\text{d}}^{'})\big|\big|p(\V{z}_{\text{d}}|\V{x})\big) + \mathbb{E}_{q(\V{z}_{\text{d}}|d;\boldsymbol{\varphi}_{\text{d}}^{'})}\big\{\log \frac{p(\V{x}|\V{z}_{\text{d}})}{p(\V{x}|d)}\big\} \\
        &\quad + \operatorname{D}_{\text{KL}}\big(q(\V{z}_{\text{e}}|k;\boldsymbol{\varphi}_{\text{e}}^{'})\big|\big|p(\V{z}_{\text{e}}|\V{x})\big) + \mathbb{E}_{q(\V{z}_{\text{e}}|k;\boldsymbol{\varphi}_{\text{e}}^{'})}\big\{\log \frac{p(\V{x}|\V{z}_{\text{e}})}{p(\V{x}|k)}\big\}
    \end{aligned}
\end{equation}

Suppose we are given a dataset $\mathcal{D}=\{\V{x}^{(n)},d^{(n)},k^{(n)}\}_{n=1}^N$ with $N$ i.i.d. data instances of signal measurements $\V{x}$, real distance $d$, and environment label $k$ for parameters learning. The ELBO and KL divergence can be approximated by the sampling average on the instances in $\mathcal{D}$, denoted as $\mathcal{L}_{\text{ELBO},\mathcal{D}}$ and $\mathcal{L}_{\text{KL},\mathcal{D}}$. The empirical ELBO can then be used as the objective function to approximate Maximum Likelihood (ML) estimation on the dataset $\mathcal{D}$ with respect to parameters $\boldsymbol{\theta}$, $\boldsymbol{\phi}$ and $\boldsymbol{\varphi}$, given as follow,
\begin{equation}  \label{eq:opt}
    \boldsymbol{\phi}^*,\boldsymbol{\theta}^*, \boldsymbol{\varphi}^* = \arg\max_{\boldsymbol{\phi},\boldsymbol{\theta}, \boldsymbol{\varphi}} \mathcal{L}_{\text{ELBO},\mathcal{D}}(\V{x},d,k;\boldsymbol{\phi},\boldsymbol{\theta}, \boldsymbol{\varphi}).
\end{equation}

We substitute $p(\V{z}_{\text{d}}|\V{x})$ with $q(\V{z}_{\text{d}}|\V{x};\boldsymbol{\phi}^*)$ and $p(\V{x}|\V{z}_{\text{d}})$ with $p(\V{x}|\V{z}_{\text{d}};\boldsymbol{\theta}^*)$ in \eqref{eq:kl_gen}. Then the optimization for parameter $\boldsymbol{\varphi}^{'}$ is given as follows,
\begin{equation}  \label{eq:opt_gen}
    \boldsymbol{\varphi}^{'*} = \arg\min_{\boldsymbol{\varphi}^{'}} \mathcal{L}_{\text{KL},\mathcal{D}}(\V{x},d,k;\boldsymbol{\phi}^*,\boldsymbol{\theta}^*,\boldsymbol{\varphi}^{'}).
\end{equation}

Once the parameters $\boldsymbol{\phi}^*$, $\boldsymbol{\theta}^*$, $\boldsymbol{\varphi}^*$ and $\boldsymbol{\varphi}^{'*}$ are obtained in the offline phase, the two types of signal generation can be achieved with corresponding distributions. The label-based signal synthesis can be achieved by sampling features via conditional distributions $q(\V{z}_{\text{d}}|d;\boldsymbol{\varphi}_{\text{d}}^{'*})$ and $q(\V{z}_{\text{e}}|k;\boldsymbol{\varphi}_{\text{e}}^{'*})$, and then sampling signal instance from the conditional distribution $p(\V{x}|\V{z}_{\text{d}},\V{z}_{\text{k}};\boldsymbol{\theta}^{*})$. In addition, the signal-based signal translation can be achieved by getting distance features by sampling $q(\V{z}_{\text{d}},\V{z}_{\text{e}}|\V{x};\boldsymbol{\phi}^{*})$, environment features by sampling $q(\V{z}_{\text{e}}|k;\boldsymbol{\varphi}_{\text{e}}^{'*})$, and then sampling signal instance from the conditional distribution $p(\V{x}|\V{z}_{\text{d}},\V{z}_{\text{k}};\boldsymbol{\theta})$. We construct VI method to approximate these unknown distributions.

In the following, we form the distribution assumptions to specify the objectives $\mathcal{L}_{\text{ELBO},\mathcal{D}}$ and $\mathcal{L}_{\text{KL},\mathcal{D}}$. We then describe the process to obtain optimal parameters $\boldsymbol{\phi}$, $\boldsymbol{\theta}$, $\boldsymbol{\varphi}$, and $\boldsymbol{\varphi}^{'}$ for the unknown distributions via data learning.

\subsection{Implicit Distribution Assumption}
\label{sec:m_imp}

In order to specify the VI method, we make further assumptions on the parametric distributions to derive the analytical/empirical objective function for network learning.

We first assume that the prior over each latent variable is modeled by an isotropic multivariate Gaussian:
\begin{equation}  \label{eq:priors}
    \begin{aligned}
        p(\V{z}_{\text{d}}) &= \mathcal{N}(\V{z}_{\text{d}}; \boldsymbol{0},\epsilon_{\text{d}}\boldsymbol{I})  \\
        p(\V{z}_{\text{e}}) &= \mathcal{N}(\V{z}_{\text{e}}; \boldsymbol{0},\epsilon_{\text{e}}\boldsymbol{I})
    \end{aligned}
\end{equation}
\noindent where $\epsilon_{\text{d}}$ and $\epsilon_{\text{e}}$ are small values arbitrarily given to interpret randomness.

We then relax the Deep Gaussian approximations in \cite{LiMazShe:J23} on $q(\V{z}_{\text{d}},\V{z}_{\text{e}}|\V{x})$ and $p(\V{x}|\V{z}_{\text{d}},\V{z}_{\text{e}})$ to more general implicit distribution approximations \cite{Fer:C17}. Such relaxation benefits for realistic signal generation since the target distribution are more complicated than the former inference tasks in \cite{LiMazShe:J23}.

\begin{figure}[!t]
\centering
\subfloat[]{\includegraphics[height=1.2in]{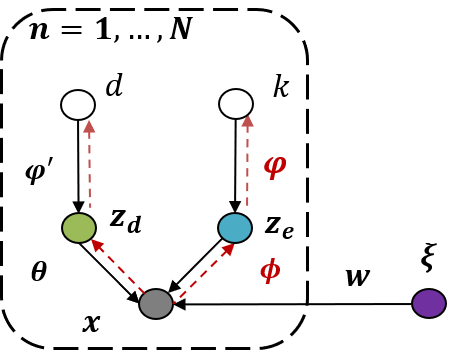}%
\label{fig:graphic_c}}
\caption{Illustration of adversarial variable $\xi$ and implicit distribution assumption. (a) Graphical model with the global adversarial variable $\xi$.
(b) Variable $\xi$ tells the truthfulness via sampling their empirical distributions from the true and generated observed variable $\V{x}$ and latent variables $\V{z}_{\text{d}}$, $\V{z}_{\text{e}}$.
}
\label{fig:graphic2}
\end{figure}

\textit{Assumption 3.} (Implicit Distribution) In the following we use the assumptions:
\begin{enumerate}
    \item[1)] We introduce a global binary variable $\xi\in\{0, 1\}$, as depicted in Fig.~\ref{fig:graphic2}.
    \item[2)] Suppose the additional binary variable is defined on the LVM variables where
    \begin{equation}  \label{eq:ratio}
        \begin{aligned}
            q(\V{z}_{\text{d}},\V{z}_{\text{e}}|\V{x};\boldsymbol{\phi},\boldsymbol{\omega})p_{\mathcal{D}}(\V{x};\boldsymbol{\phi},\boldsymbol{\omega}) &= p(\V{x}, \V{z}_{\text{d}},\V{z}_{\text{e}}|\xi=1)  \\
            p(\V{z}_{\text{d}},\V{z}_{\text{e}})p(\V{x}|\V{z}_{\text{d}},\V{z}_{\text{e}};\boldsymbol{\theta},\boldsymbol{\omega}) &= p(\V{x}, \V{z}_{\text{d}},\V{z}_{\text{e}}|\xi=0)
        \end{aligned}
    \end{equation}
\end{enumerate}

\subsection{Empirical VI Objectives}
\label{sec:m_obj}

According to the implicit assumption in \eqref{eq:ratio}, we can achieve the estimation of parameters $\boldsymbol{\omega}$ of variable $\xi$ via maximum likelihood (ML). We have the following empirical objective function over dataset $\mathcal{D}$.

\textit{Proposition 2:} (Empirical ML/Discriminative Objective) If \textit{Assumptions 3} is satisfied, we have the following empirical maximum likelihood objective for variable $\xi$ and its parameter $\boldsymbol{\omega}$ on dataset $\mathcal{D}=\{\V{x}^{(n)},d^{(n)},k^{(n)}\}_{n=1}^N$,
\begin{equation}
    \label{eq:posi_1}
    \begin{aligned}
        &\quad\mathcal{L}_{\text{ML},\mathcal{D}}(\V{x},\V{z}_{\text{d}},\V{z}_{\text{e}};\boldsymbol{\omega}) \\
        &= \mathbb{E}_{\V{x},\V{z}_{\text{d}},\V{z}_{\text{e}}}\big\{p(\V{x},\V{z}_{\text{d}},\V{z}_{\text{e}},\xi)\log q(\V{x},\V{z}_{\text{d}},\V{z}_{\text{e}},\xi;\boldsymbol{\omega}) \big\}  \\
        &= p(\xi=0)\mathbb{E}_{\V{x},\V{z}}\big\{p(\V{x},\V{z}|\xi=0)\log q(\xi=0|\V{x},\V{z};\boldsymbol{\omega}) \big\}  \\
        &= 0.5\Big(\mathbb{E}_{p_{\mathcal{D}}(\V{x})}\mathbb{E}_{q(\V{z}|\V{x};\boldsymbol{\phi})}\big\{\log q(\xi=0|\V{x},\V{z};\boldsymbol{\omega}) \big\}  \\
        &\phantom{=0.5\Big(} +\mathbb{E}_{p(\V{z})}\mathbb{E}_{p(\V{x}|\V{z};\boldsymbol{\theta})}\Big\{\log \big(1-q(\xi=0|\V{x},\V{z};\boldsymbol{\omega})\big) \Big\}\Big)
    \end{aligned}
\end{equation}
\noindent where we denote $\V{z}=\{\V{z}_{\text{d}},\V{z}_{\text{e}}\}$ for convenience.

We further have the following specification of the ELBO \eqref{eq:bound_pro} over dataset $\mathcal{D}$.

\textit{Proposition 3:} (Empirical ELBO/Feature Disentanglement Objective) If \textit{Assumptions 3} is satisfied, we have the following empirical form of the ELBO on dataset $\mathcal{D}=\{\V{x}^{(n)},d^{(n)},k^{(n)}\}_{n=1}^N$,
\begin{equation}
    \begin{aligned}
    &\quad \mathcal{L}_{\text{ELBO},\mathcal{D}}(\V{x},d,k;\boldsymbol{\phi},\boldsymbol{\theta},\boldsymbol{\varphi}) \\
    &= \mathbb{E}_{p_{\mathcal{D}}(\V{x})}\mathbb{E}_{q(\V{z}_{\text{d}},\V{z}_{\text{e}}|\V{x};\boldsymbol{\phi})}\big\{\log \frac{p(\V{x}|\V{z}_{\text{d}},\V{z}_{\text{e}};\boldsymbol{\theta})p(\V{z}_{\text{d}},\V{z}_{\text{e}})}{q(\V{z}_{\text{d}},\V{z}_{\text{e}}|\V{x};\boldsymbol{\phi})}\big\}  \\
    &\quad + \mathbb{E}_{p_{\mathcal{D}}(\V{x})}\mathbb{E}_{q(\V{z}_{\text{d}}|\V{x};\boldsymbol{\phi})}\big\{\log p(d|\V{z}_{\text{d}};\boldsymbol{\varphi})\big\}  \\
    &\quad + \mathbb{E}_{p_{\mathcal{D}}(\V{x})}\mathbb{E}_{q(\V{z}_{\text{e}}|\V{x};\boldsymbol{\phi})}\big\{\log p(k|\V{z}_{\text{e}};\boldsymbol{\varphi})\big\}  \\
    &\leq \log~p(\V{x}, d, k)
    \end{aligned}
\end{equation}

With the adversarial variable $\boldsymbol{\xi}$ and implicit distribution assumption, the bound can be further expressed in an analytical form as follows:
\begin{equation}  \label{eq:posi_2}
    \begin{aligned}
    &\quad \mathcal{L}_{\text{ELBO},\mathcal{D}}(\V{x},d,k;\boldsymbol{\phi},\boldsymbol{\theta},\boldsymbol{\varphi}) \\
    &=:  \mathbb{E}_{p_{\mathcal{D}}(\V{x})}\mathbb{E}_{q(\V{z}_{\text{d}},\V{z}_{\text{e}}|\V{x};\boldsymbol{\phi})}\big\{\log \frac{q(\xi=0|\V{x},\V{z}_{\text{d}},\V{z}_{\text{e}};\boldsymbol{\omega})}{1-q(\xi=0|\V{x},\V{z}_{\text{d}},\V{z}_{\text{e}};\boldsymbol{\omega})}\big\}  \\
    &\quad + \mathbb{E}_{p_{\mathcal{D}}(\V{x})}\mathbb{E}_{q(\V{z}_{\text{d}}|\V{x};\boldsymbol{\phi})}\big\{\log p(d|\V{z}_{\text{d}};\boldsymbol{\varphi})\big\}  \\
    &\quad + \mathbb{E}_{p_{\mathcal{D}}(\V{x})}\mathbb{E}_{q(\V{z}_{\text{e}}|\V{x};\boldsymbol{\phi})}\big\{\log p(k|\V{z}_{\text{e}};\boldsymbol{\varphi})\big\}
    \end{aligned}
\end{equation}

The empirical KL divergence $\mathcal{L}_{\text{KL},\mathcal{D}}$ for the feature generation objective from \textit{Proposition 1} can be approximated by Mean Square Error between data points.
Hence, the inference on LVM is conducted by addressing the optimization $\max_{\boldsymbol{\omega}} \mathbb{L}_{\text{ML},\mathcal{D}}(\V{x},\V{z}_{\text{d}},\V{z}_{\text{e}}; \boldsymbol{\omega})$,  $\max_{\boldsymbol{\phi},\boldsymbol{\theta},\boldsymbol{\varphi}} \mathbb{L}_{\text{ELBO},\mathcal{D}}(\V{x},d,k; \boldsymbol{\phi},\boldsymbol{\theta},\boldsymbol{\varphi})$, and $\min_{\boldsymbol{\varphi}^{'}} \mathbb{L}_{\text{KL},\mathcal{D}}(\V{x},d,k; \boldsymbol{\varphi}^{'})$ by means of stochastic gradient descent algorithm.

\section{IIns-GAN Network Implementation}
\label{sec:network}

This section proposes a DL network, namely the Inter-Instance GAN (IIns-GAN), to implement the realistic signal generation framework. We first introduce the network structure and the loss function derived from the empirical VI objectives for network training. Then the DL algorithms of offline training and online testing are presented.

\subsection{GAN-Based Network Structure}
\label{sec:net_struc}

\begin{figure*}[tbp]
  \centering
  \includegraphics[width=6in]{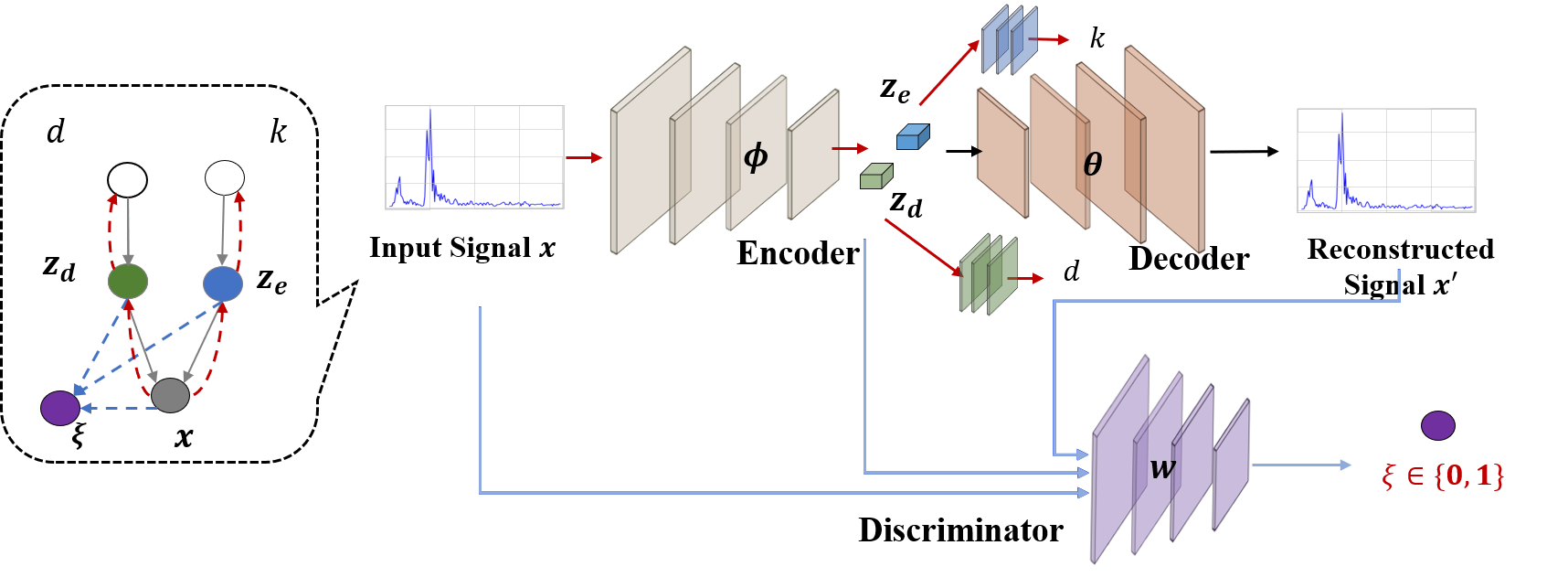}\\
  \centering
  \vspace{-0.3cm}
  \caption{Network structure of the proposed IIns-GAN. The network parameters are trained on empirical VI objectives  in the proposed LVM.}  
  \label{fig:structure}
  \vspace{-0.3cm}
\end{figure*}

The VI method is targeted to approximate the variational posterior distributions $q(\V{z}_{\text{d}}|d;\boldsymbol{\varphi}_{\text{d}}^{'})$, $q(\V{z}_{\text{e}}|k;\boldsymbol{\varphi}_{\text{e}}^{'})$ and $q(\V{z}_{\text{d}},\V{z}_{\text{e}}|\V{x};\boldsymbol{\phi})$. Based on the assumptions, the method is required to further estimate the intractable conditional distributions for likelihoods $p(\V{x}|\V{z}_{\text{d}},\V{z}_{\text{e}};\boldsymbol{\theta})$, $p(d|\V{z}_{\text{d}};\boldsymbol{\varphi}_{\text{d}})$, and $p(k|\V{z}_{\text{e}};\boldsymbol{\varphi}_{\text{e}})$, as well as the discriminative distribution $p(\xi|\V{x}, \V{z}_{\text{d}}, \V{z}_{\text{e}};\boldsymbol{\omega})$.

The network structure is presented in Fig.~\ref{fig:structure}. The overall structure consists of three main modules: an IIns-VAE with network parameters $\{\boldsymbol{\phi}, \boldsymbol{\theta}, \boldsymbol{\varphi}\}$, an inverse estimator with parameter $\boldsymbol{\varphi}^{'}$, and a discriminator with parameter $\boldsymbol{\omega}$. The three modules are trained in parallel with three alternative flows, following the strategy of GANs \cite{Goodfellow2014GenerativeAN}.

In the online phase for practical use, distribution parameters are fixed and the two types of signal generation can be achieved with corresponding distributions. Signals are generated via either the inverse estimator or the encoder together with the decoder, instead of utilizing the whole framework.
Specifically, the label-based signal synthesis is conducted by first estimating distance and environment features $\V{z}_{\text{d}}$ from the given label $d$ via the inverse estimator, achieving  and $\V{z}_{\text{e}}$ from label $k$ via a saved mapping, and then generating target signal $\V{x}_{d,k}$. The signal-based translation is conducted by first disentangling the distance feature $\V{z}_{\text{d}}$ from the given source signal $\V{x}_{k'}$ via the encoder, achieving the environment feature $\V{z}_{\text{e}}$ from the given label $k$ via the saved mapping, and then generating target signal via the decoder.

Note that the general framework proposed for variational learning can be utilized with general network structures. In particular, the approach proposed could also be implemented using other types of neural networks including residual network (ResNet) and long-short term memory (LSTM) networks.
The implementation details are further listed in Section \ref{sec:e_net}.

\subsection{Loss Functions from Empirical VI Objectives}
\label{sec:net_obj}

We construct the discriminator network to learn the additional binary variable $\xi\in\{0, 1\}$. Suppose a discriminator network with parameters $\boldsymbol{\omega}$ learns the function $h(\cdot,\cdot;\boldsymbol{\omega}):\V{x}, \V{z}_{\text{d}},\V{z}_{\text{e}} \to \xi$. We assume that $h(\V{x},\V{z};\boldsymbol{\omega})=q(\xi=0|\V{x},\V{z};\boldsymbol{\omega})$.
We denote $\V{z}=\{\V{z}_{\text{d}}, \V{z}_{\text{e}}\}$ for convenience. The loss function for the discriminator network, derived from the empirical maximum likelihood \eqref{eq:posi_1} and \textit{Assumption 3} \eqref{eq:ratio}, can be constructed as follows,
\begin{equation}  \label{eq:l_dis}
    \begin{aligned}
        \mathbb{L}_{\text{dis}}(\mathcal{D};\boldsymbol{\omega}) &= \mathbb{E}_{ p_{\mathcal{D}}(\V{x})}\mathbb{E}_{ q(\V{z}|\V{x};\boldsymbol{\phi}))} \log \big(h(\V{x},\V{z};\boldsymbol{\omega})\big)  \\
        &\quad + \mathbb{E}_{ p(\V{z})}\mathbb{E}_{p(\V{x}|\V{z};\boldsymbol{\theta})} \big(1-\log h(\V{x},\V{z};\boldsymbol{\omega})\big)
    \end{aligned}
\end{equation}
\noindent where the learnable parameter is only $\boldsymbol{\omega}$, with the other parameters $\boldsymbol{\phi}$ and $\boldsymbol{\theta}$ only present to create the distributions.

The IIns-VAE network learns the encoder function $g(\cdot;\boldsymbol{\phi}):\V{x} \to \V{z}_{\text{d}},\V{z}_{\text{e}}$, the decoder function $g(\cdot;\boldsymbol{\theta}):\V{z}_{\text{d}},\V{z}_{\text{e}} \to \V{x}$, and two regularization functions $f(\cdot;\boldsymbol{\phi}_{\text{d}}): \V{z}_{\text{d}} \to d$, $f(\cdot;\boldsymbol{\phi}_{\text{e}}): \V{z}_{\text{e}} \to k$. The loss function for the IIns-VAE, derived from the negative empirical ELBO \eqref{eq:posi_2} and \textit{Assumption 3} \eqref{eq:ratio}, can be constructed as follows,
\begin{equation}  \label{eq:l_ae}
    \begin{aligned}
    \mathbb{L}_{\text{vae}}(\mathcal{D};\boldsymbol{\phi},\boldsymbol{\theta},\boldsymbol{\varphi}) &= \mathbb{E}_{p_{\mathcal{D}}(\V{x})}\mathbb{E}_{q(\V{z}|\V{x};\boldsymbol{\phi})} \log \frac{1-h(\V{x},\V{z};\boldsymbol{\omega})}{h(\V{x},\V{z};\boldsymbol{\omega})} \\
    &\quad + \mathbb{E}_{p_{\mathcal{D}}(\V{x},d)}\big\Vert d-\hat{d}(\V{x};\boldsymbol{\phi},\boldsymbol{\varphi}_{\text{d}})\big\Vert^2 \\
    &\quad -\mathbb{E}_{p_{\mathcal{D}}(\V{x},k)}\log q(k|\V{x};\boldsymbol{\phi},\boldsymbol{\varphi}_{\text{e}})
    \end{aligned}
\end{equation}
\noindent where the learnable parameters are $\boldsymbol{\phi}$, $\boldsymbol{\theta}$ and $\boldsymbol{\varphi}$. $\hat{d}(\V{x};\boldsymbol{\phi},\boldsymbol{\varphi}_{\text{d}})$ denotes the distance estimated by the encoder and the estimator network with parameters $\boldsymbol{\phi}$ and $\boldsymbol{\varphi}_{\text{e}}$.

The third flow with   with the inverse estimator and decoder network learn the feature generation functions $f^{'}(\cdot;\boldsymbol{\varphi}_{\text{d}}^{'}): d \to \V{z}_{\text{d}}$, $f^{'}(\cdot;\boldsymbol{\varphi}_{\text{e}}^{'}): k \to \V{z}_{\text{e}}$, and the decoder function $g(\cdot;\boldsymbol{\theta}):\V{z}_{\text{d}},\V{z}_{\text{e}} \to \V{x}$.
Since $k$ and the corresponding $\V{z}_{\text{e}}$ are discrete variables, the mapping $f^{'}(\cdot;\boldsymbol{\varphi}_{\text{e}}^{'})$ from labels to environment features are saved during learning as a dictionary instead of learning a separate 'inverse classifier' network.
The empirical loss term for the third flow, derived from the empirical KL objective \eqref{eq:kl_gen}, can be constructed as follows,
\begin{equation}  \label{eq:l_inv}
    \begin{aligned}
    \mathbb{L}_{\text{inv}}(\mathcal{D};\boldsymbol{\varphi}^{'},\boldsymbol{\theta}) &= \mathbb{E}_{p_{\mathcal{D}}(\V{x})}\big\Vert \V{z}_{\text{d}}-\hat{\V{z}}_{\text{d}}(d;\boldsymbol{\varphi}^{'})\big\Vert^2 \\
    &\quad
+ \mathbb{E}_{p_{\mathcal{D}}(\V{x})}\big\Vert \V{x}-\hat{\V{x}}(\hat{\V{z}}_{\text{d}};\boldsymbol{\theta})\big\Vert^2
    \end{aligned}
\end{equation}
\noindent where the learnable parameter are $\boldsymbol{\varphi}^{'}$ and $\boldsymbol{\theta}$.

Hence, the estimation of intractable distributions on LVM is conducted by addressing the optimizations on dataset $\mathcal{D}$ by means of stochastic gradient descent algorithm.

\subsection{Deep Learning Algorithms}
\label{sec:net_alg}

\begin{figure}[!t]
\centering
\qquad  
{\includegraphics[height=2.4in]{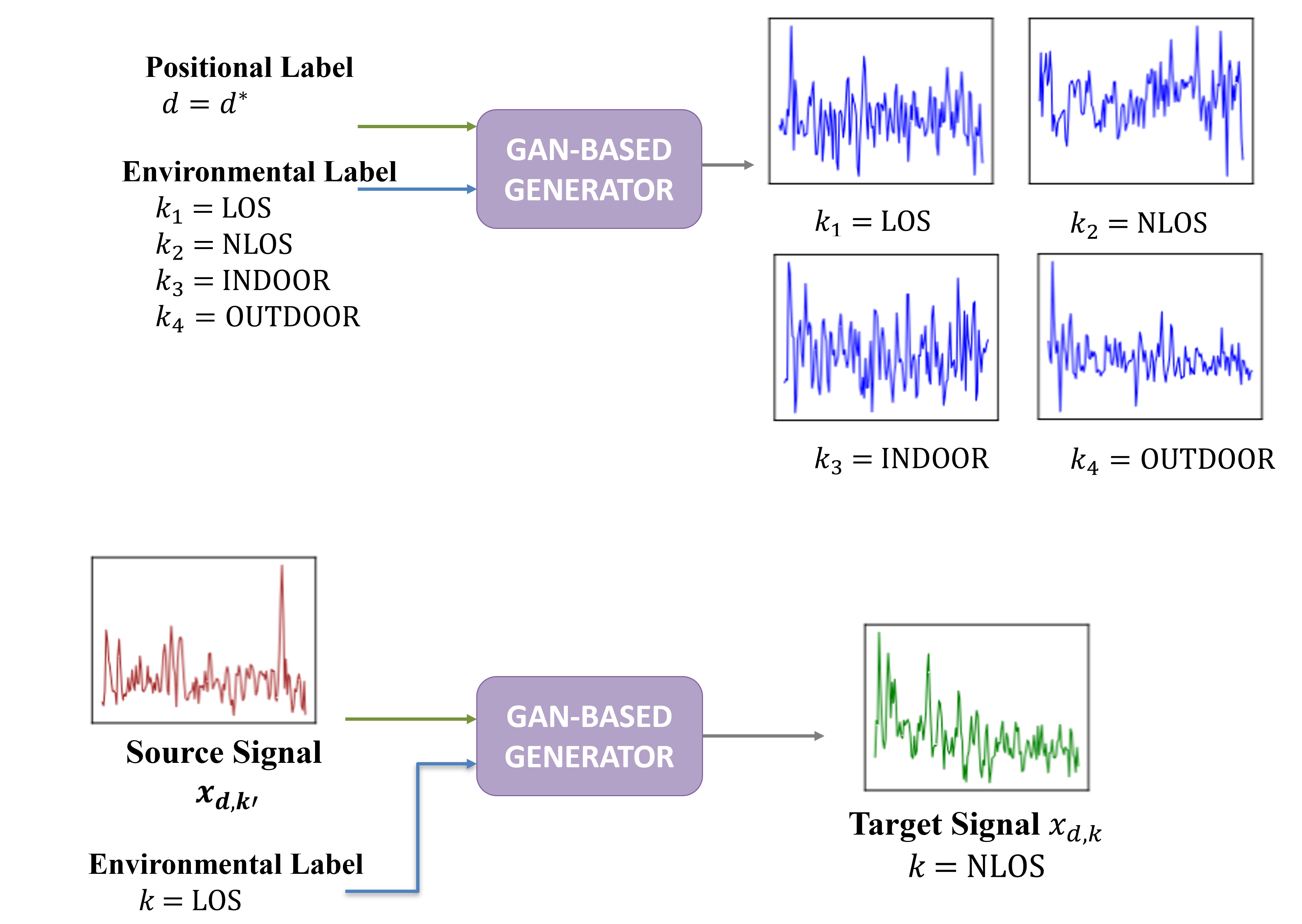}}%

\caption{Illustration of the methodology and usage of two types of realistic signal generations.
The upper graph shows the label-based synthesis and the lower graph shows the signal-based translation.}
\label{fig:use_intro}
\end{figure}

The offline training phase includes three alternative training flows with respect to the three network modules. On one flow, IIns-VAE takes in signal measurements $\V{x}$, disentangles features $\V{z}_{\text{d}}$ and $\V{z}_{\text{e}}$ via $\boldsymbol{\phi}$, and produces the reconstructed $\V{x}$ via network parameters $\boldsymbol{\theta}$, i.e., $\V{z}_{\text{d}},\V{z}_{\text{e}}\sim q(\V{z}_{\text{d}},\V{z}_{\text{e}}|\V{x};\boldsymbol{\phi})$, $\V{\hat{x}}\sim p(\V{x}|\V{z}_{\text{d}},\V{z}_{\text{e}};\boldsymbol{\theta})$. Note that the estimator and identifier in IIns-VAE serve as regularization to help feature disentanglement. On the second flow, the inverse estimator net takes in distance label $d$ and generate distance features $\V{z}_{\text{d}}$ with parameters $\boldsymbol{\varphi}^{'}_{\text{d}}$, which is further fed into the decoder for a generated signal $\hat{\V{x}}^{'}$, i.e., $\hat{\V{z}}_{\text{d}}\sim q(\V{z}_{\text{d}}|d;\boldsymbol{\varphi}^{'}_{\text{d}})$, $\hat{\V{x}}^{'}\sim p(\V{x}|{\V{\hat{z}}}_{\text{d}},\V{z}_{\text{e}};\boldsymbol{\theta})$. On the third flow, the discriminator network takes in the real signal $\V{x}$, reconstructed signal $\V{\hat{x}}$ and generated signal $\V{\hat{x}}^{'}$ while producing the classification result $\xi$ to tell their soundness. The network learn these parameters with loss functions derived from the empirical VI objectives to guide training.

In the online phase for practical use, distribution parameters $\boldsymbol{\phi}^*$, $\boldsymbol{\theta}^*$, $\boldsymbol{\varphi}^*$ and $\boldsymbol{\varphi}^{'*}$ are fixed and generated signals can be achieved with corresponding distributions. 
We show two typical types of signal generation as case studies: 1) Label-based synthesis and 2) signal-based translation.

The label-based generation for signal synthesis is conducted by the following two steps:
\begin{enumerate}
    \item[1)] Estimate distance and environment features $\V{z}_{\text{d}}$ and $\V{z}_{\text{e}}$ from given labels $d$ and $k$ based on $q(\V{z}_{\text{d}}|\V{x};\boldsymbol{\varphi}_{\text{d}}^{'*})$ and $q(\V{z}_{\text{e}}|\V{x};\boldsymbol{\varphi}_{\text{e}}^{'*})$, respectively.
    \item[2)] Generate target signal $\V{x}_{d,k}$ based on $p(\V{x}|\V{z}_{\text{d}},\V{z}_{\text{e}};\boldsymbol{\theta}^*)$ with distance and environment features $\V{z}_{\text{d}}$ and $\V{z}_{\text{e}}$.
\end{enumerate}

The signal-based translation for signal synthesis is conducted by the following three steps:
\begin{enumerate}
    \item[1)] Estimate the distance feature $\V{z}_{\text{d}}$ from the given source signal $\V{x}_{k'}$ based on $q(\V{z}_{\text{d}},\V{z}_{\text{e}}|\V{x};\boldsymbol{\phi}^*)$.
    \item[2)] Estimate the environment feature $\V{z}_{\text{e}}$ from the given label $k$ based on $q(\V{z}_{\text{e}}|\V{x};\boldsymbol{\varphi}_{\text{e}}^{'*})$, respectively.
    \item[3)] Generate target signal $\V{x}_{d,k}$ based on $p(\V{x}|\V{z}_{\text{d}},\V{z}_{\text{e}};\boldsymbol{\varphi}^*)$ with distance and environment features $\V{z}_{\text{d}}$ and $\V{z}_{\text{e}}$.
\end{enumerate}

Algorithm \ref{alg:iins-gan-off}-\ref{alg:iins-gan-on} outlines the online and offline phases for training and utilizing IIns-GAN for the signal generation tasks.
Further downstream tasks potential to be enabled by IIns-GAN are discussed in Sec.~\ref{sec:d_var}.

\begin{algorithm}[!t]
  \caption{Variational Learning of IIns-GAN}
  \label{alg:iins-gan-off}
  \textbf{Offline Phase}
    \begin{algorithmic}[1]
    \Require Training dataset $\mathcal{D}=\{\V{x}^{(n)},{d}^{(n)},k\}_{n=1}^{N}$, training iterations $TK$, learning rate $\alpha$, batch size $m$, initial network parameters $\boldsymbol{\phi}_0$, $\boldsymbol{\theta}_0$, $\boldsymbol{\varphi}_0$, $\boldsymbol{\varphi}'_0$, and $\boldsymbol{\omega}_0$.
    \Ensure Optimized parameters $\boldsymbol{\phi}^{*}$, $\boldsymbol{\theta}^{*}$, $\boldsymbol{\varphi}^{*}$, $\boldsymbol{\varphi}'^{*}$, and $\boldsymbol{\omega}^{*}$.
    \For {0$<$n$<$$TK$}
        \State Sample a minibatch $\mathcal{B}=\{\V{x}^{(n)},d^{(n)},k^{(n)}\}_{n=1}^{m} \sim \mathcal{D}$.
        \State Update $\boldsymbol{\phi}$, $\boldsymbol{\theta}$, $\boldsymbol{\varphi}^{*}$ by ascending stochastic gradient: $$\nabla\mathbb{L}(\mathcal{B};\boldsymbol{\phi}, \boldsymbol{\theta},\boldsymbol{\varphi})$$
        \State Sample a minibatch $\mathcal{B}=\{\V{x}^{(n)},d^{(n)},k^{(n)}\}_{n=1}^{m} \sim \mathcal{D}$.
        \State Update $\boldsymbol{\varphi}^{'}$ by ascending stochastic gradient: $$\nabla\mathbb{L}(\mathcal{B};\boldsymbol{\varphi}^{'})$$
        \State Sample a minibatch $\mathcal{B}=\{\V{x}^{(n)},d^{(n)},k^{(n)}\}_{n=1}^{m} \sim \mathcal{D}$.
        \State Generate reconstructed signals via parameters $\boldsymbol{\phi}$, $\boldsymbol{\theta}$
        $$\mathcal{\hat{B}}=\{\V{\hat{x}}^{(n)}\}_{n=1}^{m}$$
        \State Generate synthesized signals via parameters $\boldsymbol{\varphi}^{'}$, $\boldsymbol{\theta}$:
        $$\mathcal{\hat{B}}^{'}=\{\V{\hat{x}}^{'(n)}\}_{n=1}^{m}$$
        \State Update $\boldsymbol{\omega}$ by ascending stochastic gradient: $$\nabla\mathbb{L}(\mathcal{B},\mathcal{\hat{B}},\mathcal{\hat{B}}^{'};\boldsymbol{\omega})$$.
    \EndFor
  \end{algorithmic}
\end{algorithm}

 \begin{algorithm}[!t]
  \caption{Variational Learning of IIns-GAN}
  \label{alg:iins-gan-on}
  \textbf{Online Phase (Case Study 1: Label-Based Synthesis)}
    \begin{algorithmic}[1]
    \Require Distance label ${d}^{(n)}$, environment label $k^{(h)}$, the stored inverse regressor and classifier parameters $\boldsymbol{\varphi}'^*=\{\boldsymbol{\varphi}'^{*}_{d}, \boldsymbol{\varphi}'^{*}_{k}\}$, decoder parameters  $\boldsymbol{\theta}^{*}$.
    \Ensure Generated signal $\V{x}_{d,k}^{(n)}$.
    \For{$n\geq 0$}
    \State Generate $\V{z}_{\text{d}}^{(n)}$, $\V{z}_{\text{e}}^{(n)}$ via parameters $\boldsymbol{\varphi}^{'*}_{d}$,$\boldsymbol{\varphi}^{'*}_{k}$.
\begin{equation*}
    \begin{aligned}
    \V{z}_{\text{d}}^{(n)}\sim & \mathcal{N}\big(\V{z}_{\text{d}};\boldsymbol{\mu}(\V{x}^{(n)},d^{(n)};\boldsymbol{\varphi}'^{*}_{d});\boldsymbol{\sigma}^2(\V{x}^{(n)},d^{(n)};\boldsymbol{\varphi}'^{*}_{d})\mathbf{I}\big)  \\
    \V{z}_{\text{e}}^{(n)}\sim & \mathcal{N}\big(\V{z}_{\text{e}};\boldsymbol{\mu}(\V{x}^{(n)},k^{(n)};\boldsymbol{\varphi}'^{*}_{k});\boldsymbol{\sigma}^2(\V{x}^{(n)},k^{(n)};\boldsymbol{\varphi}'^{*}_{k})\mathbf{I}\big)
    \end{aligned}
\end{equation*}
    \State Generate $\V{x}^{(n)}_{d,k}$ via decoder parameters $\boldsymbol{\theta}^*$.
\begin{equation*}
    \V{x}^{(n)}_{d,k} \sim p(\V{x}|\V{z}_{\text{d}, \V{z}_{\text{e}}};\boldsymbol{\theta})
\end{equation*}
    \EndFor
  \end{algorithmic}

\rule[0.25\baselineskip]{
    0.48\textwidth}{
    0.5pt} \\  
  \textbf{Online Phase (Case Study 2: Signal-Based translation)}
    \begin{algorithmic}[1]
    \Require New signal measurements $\V{x}_{d,k'}^{(n)}$, environment label $k$, the stored encoder parameters $\boldsymbol{\phi}^*$, inverse regressor parameters $\boldsymbol{\varphi}'^{*}_{d}$, decoder parameters $\boldsymbol{\theta}^*$.
    \Ensure Generated signal $\V{x}_{d,k}^{(n)}$.
    \For{$n\geq 0$}
    \State Disentangle $\V{z}_{\text{e}}^{(n)}$ from $\V{x}^{(n)}$ via parameters $\boldsymbol{\phi}^*$.
\begin{equation*}
\V{z}_{\text{e}}^{(n)}\sim \mathcal{N}\big(\V{z}_{\text{e}};\boldsymbol{\mu}(\V{x}^{(n)};\boldsymbol{\phi}^*);\boldsymbol{\sigma}^2(\V{x}^{(n)};\boldsymbol{\phi}^*)\mathbf{I}\big)
\end{equation*}
    \State Generate $\V{z}_{\text{d}}^{(n)}$ from label ${d}^{(n)}$ via parameters $\boldsymbol{\varphi}'^*_{d}$.
\begin{equation*}
    \V{z}_{\text{d}}^{(n)}\sim \mathcal{N}\big(\V{z}_{\text{d}};\boldsymbol{\mu}(\V{x}^{(n)},d^{(n)};\boldsymbol{\varphi}'^*_{d});\boldsymbol{\sigma}^2(\V{x}^{(n)},d^{(n)};\boldsymbol{\varphi}'^*_{d})\mathbf{I}\big)
\end{equation*}
    \State Generate $\V{x}^{(n)}$ via decoder parameter $\boldsymbol{\theta}^*$.
\begin{equation*}
    \V{x}^{(n)} \sim p(\V{x}|\V{z}_{\text{d}},\V{z}_{\text{e}};\boldsymbol{\theta})
\end{equation*}
    \EndFor
  \end{algorithmic}
\end{algorithm}

\section{Discussion}
\label{sec:discussion}

In this section, we explore the further applications and implications of the proposed IIns-GAN framework. In particular, we discuss its versatility in various downstream tasks besides labeled signal generation, including adversarial attack detection, distance estimation, and environment identification. We then present its foundational connection to IIns-VAE \cite{LiMazShe:J23}, and the broader insights it provides for combining variational inference and DL techniques.

\subsection{Detect Signal Adversaries and Attacks}
\label{sec:d_var}

Beyond signal generation, the proposed IIns-GAN can conduct adversarial attack detection of signals, a growing concern in the field of wireless communication. Specifically, the altogether encoder and discriminator sub-structure can be used as a fake signal detector. With their parameters $\boldsymbol{\phi}$ and $\boldsymbol{\omega}$ learned and fixed, the structure can tell if an input labeled signal instance is real or fake.

The proposed IIns-GAN can enable additional downstream tasks including distance estimation and environment identification. These two tasks are enabled by the sub-structures composed of the encoder combining with the distance estimator, and with the environment identifier, respectively. With fixed parameters $\boldsymbol{\phi}$ and $\boldsymbol{\varphi}_{\text{d}}$, the estimator sub-structure can produce distance estimation from input signals. With fixed parameters $\boldsymbol{\phi}$ and $\boldsymbol{\varphi}_{\text{e}}$, the classifier sub-structure can produce environment identification from input signals.
These downstream tasks underscore the potential of IIns-GAN as a comprehensive tool for wireless signal processing. Such flexibility and scalability is critical to the developing wireless communications systems like 5G and 6G.

\subsection{Connections to IIns-VAE}
\label{sec:dis_vae}


IIns-GAN is designed based on IIns-VAE \cite{LiMazShe:J23}, a DL-based method for concurrent distance estimation and environment identification. Both methods share the same fundamental LVM and variational objective. However, IIns-GAN introduces an improved LVM with implicit distribution assumption, and proposes a GAN-based network structure to implement the generation problem. The original IIns-VAE structure is used as the generator in the proposed GAN-based network. As a result, IIns-GAN learns more precise LVM distributions and enables additional signal generation tasks.

Theoretically, the proposed IIns-GAN describe signals using the same basic LVM as IIns-VAE, which includes observed variables for signals and latent variables for distance and environment features. However, they differ in their distribution assumptions. IIns-VAE assume Deep Gaussian distributions for intractable signal distributions, while IIns-GAN relaxes this constraint by introducing an additional variable $\xi$. This discriminating variable $\xi$ allows IIns-GAN to compare real and generated signals based on their empirical distributions, providing a more flexible and capable learning model. This additional variable is further implemented by the discriminator structure in the proposed network.

Technically, IIns-GAN extends the capabilities of IIns-VAE by enabling realistic signal generation. While the decoder in IIns-VAE is used for producing regularization terms during offline training, the decoder in IIns-GAN serves as a signal generator during the online phase. This is made possible by the implicit distribution assumption and the improved GAN-shaped structure, which enhance the framework's ability to model complex distributions.

In summary, IIns-GAN improves upon IIns-VAE by integrating a discriminator to approximate more complex and intractable empirical distributions. Like IIns-VAE, the proposed IIns-GAN can be applied to signal measurements related to any positional metrics, such as angle, velocity, and acceleration, in addition to distance-related measurements in the manuscript.

\subsection{Insights for Combining Inference and Deep Learning}

The proposed IIns-GAN demonstrates the potential of integrating DL and VI for complex signal processing problems, particularly in the field of wireless sensing. This method addresses the challenge of inference with intractable variables, which conventional statistical inference methods struggle with.
In particular, the proposed IIns-GAN generates realistic signal distributions that are difficult to achieve with conventional model assumptions. This method enhances the flexibility and capability of model assumptions using network parameters and empirical data. In addition, the approach models the signal generation process in a unified Bayesian framework with interpretable objective functions, enjoying additional flexibility and transparent interpretation from statistic techniques.
The methodology of combinting VI and DL techniques opens a promising avenue for advancing signal processing and wireless communication research.

\section{Experiments}
\label{sec:exp}

In this section, we present the results of our experiments to evaluate the effectiveness of the proposed IIns-GAN. In particular, we demonstrate its capability in generating realistic wireless signals via waveform and core physical features comparison to the real signals. We further illustrate the utility of IIns-GAN in data augmentation to train learning-based models. 
Note that we utilize UWB measurements in this section, while the proposed approach is technology-agnostic and applicable to any technology providing wireless propagated signals.

\subsection{Dataset Assignment}
\label{sec:e_data}

We first describe the datasets used for training and testing our framework, as well as the details of the network implementation.

For our experiments, we utilized \textit{Dataset 3} from the IIns-VAE paper \cite{LiMazShe:J23}. This dataset, derived from a publicly available UWB database \cite{zenodo}, contains $21,250$ instances of signal measurements. Each signal measurement is associated with a real distance and environment label array. The signal measurements per sample include a waveform of length $157$ and an estimated distance from the device. The environment label array includes two labels, one for LoS or NLoS conditions and the other for indoor or outdoor scenarios.

The database was created using DecaWave EVB1000 devices and the measurements were obtained using the two-way ranging (TWR) method equipped on these commercial sensing devices. This method, enabled by IEEE 802.15.4-2011 for low-rate network devices \cite{IEEE:standard}, provides cost-effective distance estimates as it does not require synchronization devices.

We divided the dataset into a general training set, a toy-model training set, and a testing set at the partition $8:1:1$. Specifically, $80\%$ of the data was used for training the IIns-GAN, $10\%$ was used as real data for training toy models (to compare the toy models trained on generated data), and the remaining $10\%$ for testing. This setup ensures that the real data used in the data augmentation experiments was not exposed to the other toy models trained on the IIns-GAN generated data. 

While the dataset provides a case study for environment scenarios, more detailed labels for different environments can be utilized depending on the requirements of practical tasks. For instance, \textit{Dataset 4} in \cite{LiMazShe:J23} includes environment labels for rooms and blocking obstacles. All signals were standardized to have zero mean and unit variance before being fed into the network.

\subsection{Network Implementation}
\label{sec:e_net}

Nevertheless, specific labels for environment identification can be designed with regard to different requirements of practical tasks.
Loading the training dataset and standardize the signals to have zero mean and unit variance.

The network structure is composed of an IIns-VAE generator and a discriminator. The generator includes a variational encoder for feature extraction, a decoder for generation, an estimator and an identifier for regularization, and an inverse estimator for distance feature generation. The structure of the generator mirrors that of the IIns-VAE, with the inverse estimator designed to have a symmetric layer structure to the estimator. The discriminator network, on the other hand, is tasked with classifying real and generated signals given a set of signals and their corresponding labels. This network is designed as a PatchGAN classifier \cite{Isola2017ImagetoImageTW}, which aims to effectively distinguish between real and generated signals, thereby enhancing the realism of the generated signals. The specific structure consists of $3$ convolutional layers, each followed by Instance Normalization (IN) and LeakyReLU activation.

The networks were trained using the Adam optimizer \cite{DieJim:C14} with a learning rate of $0.0001$ and a batch size of $256$. The decays of first and second momentum of gradients are $0.5$ and $0.999$, respectively. The training process continued for $500$ epochs, with the model's performance evaluated at the end of each epoch. The evaluation metrics used include the RMSE and MAE between the original and generated signals, and the accuracy of the discriminator in distinguishing between real and generated signals.
The model is implemented using the Pytorch \cite{Paszke2017AutomaticDI} library and conduct learning on a GTX $1080$ GPU with a memory of $12$ GB with the accelerator powered by the NVIDA Pascal architecture. The code will be released to public in our final version.

In the following subsections, we present the results of our experiments conducted using this setup.



\subsection{Comprehensive Comparison}
\label{sec:e_signal}

\begin{figure*}[!t]
\centering
\subfloat[Real LoS signal]{\includegraphics[height=1.3in]{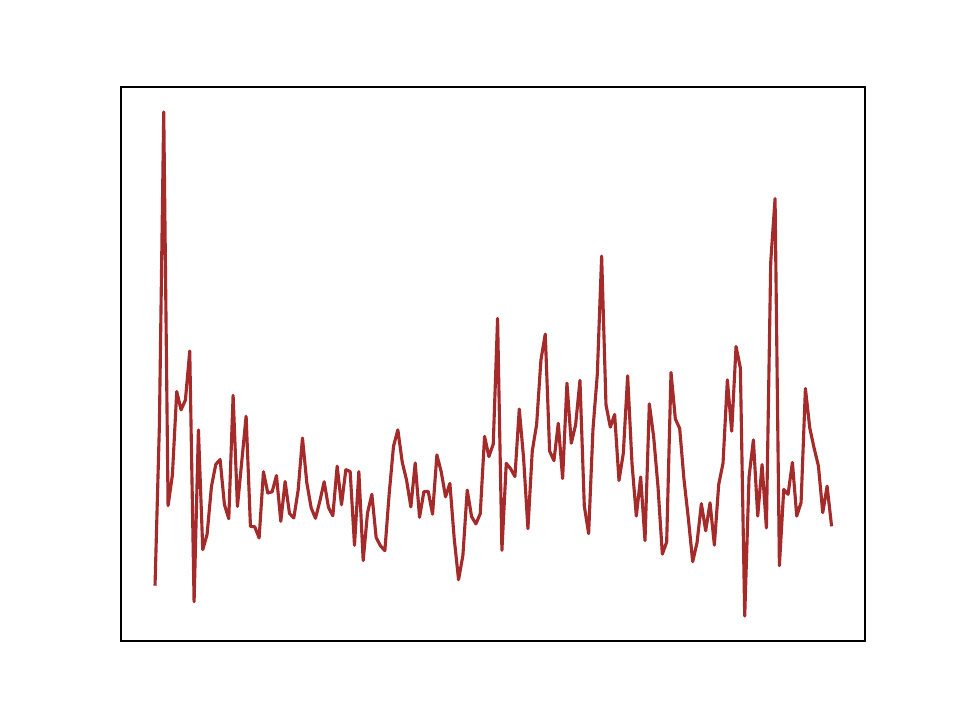}
\label{fig:cir_real_a}}
\subfloat[Real NLoS signal]{\includegraphics[height=1.3in]{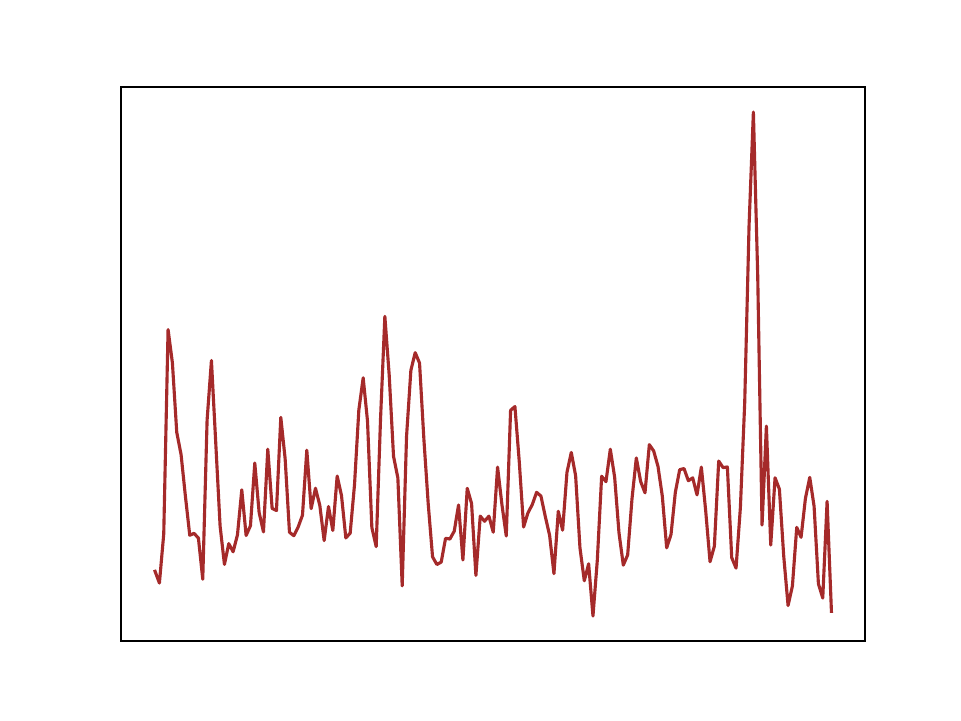}
\label{fig:cir_real_b}}
\subfloat[Real indoor signal]{\includegraphics[height=1.3in]{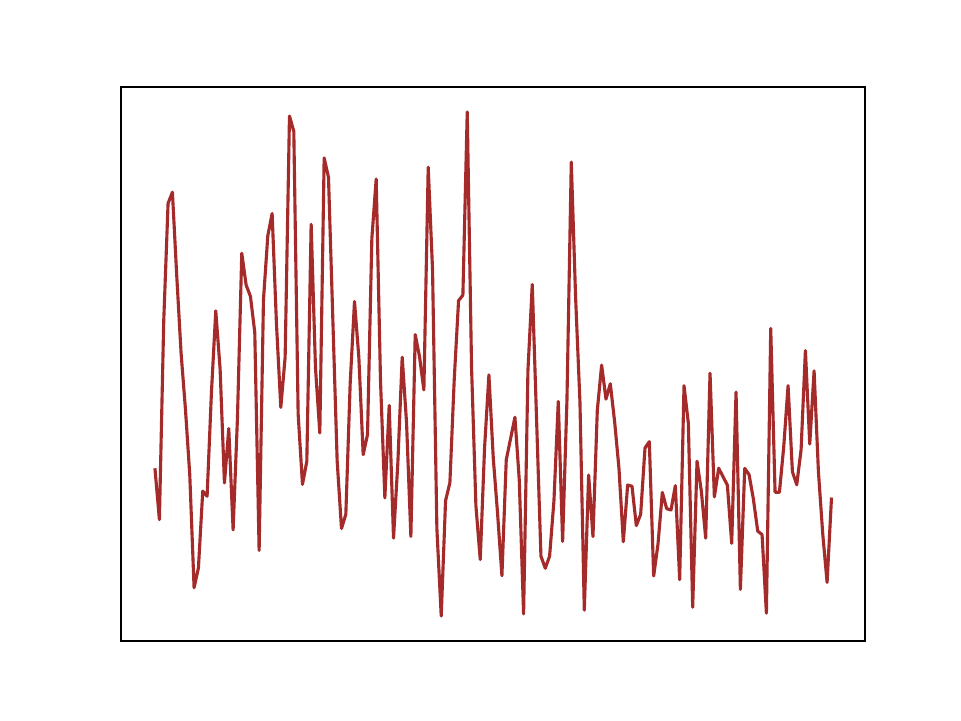}
\label{fig:cir_real_c}}
\subfloat[Real outdoor signal]{\includegraphics[height=1.3in]{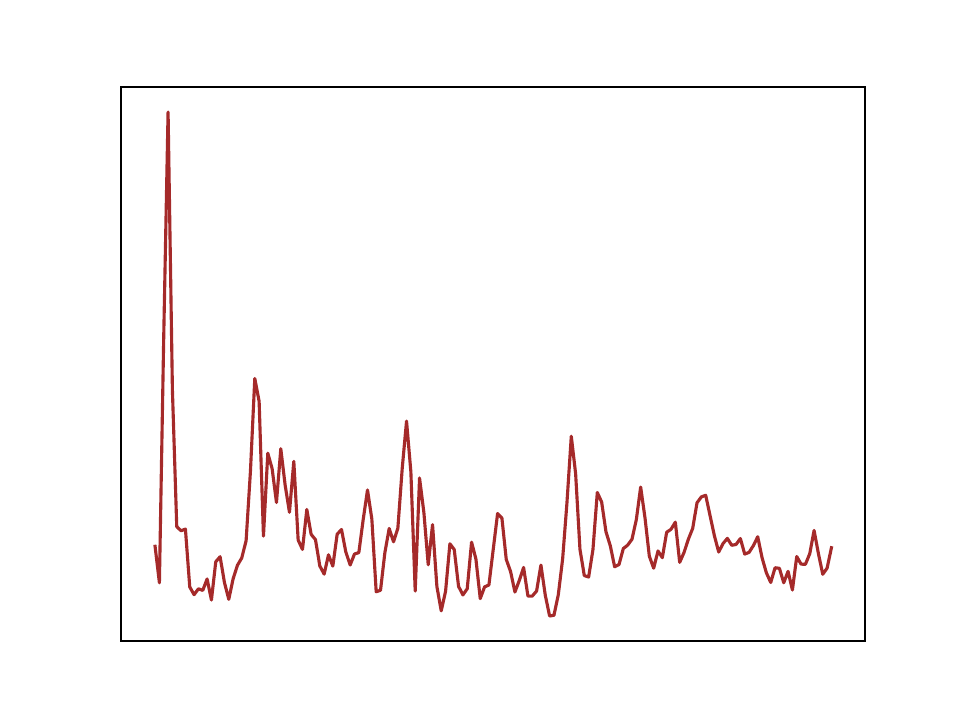}
\label{fig:cir_real_d}} \\
\subfloat[Synthesized LoS signal]{\includegraphics[height=1.3in]{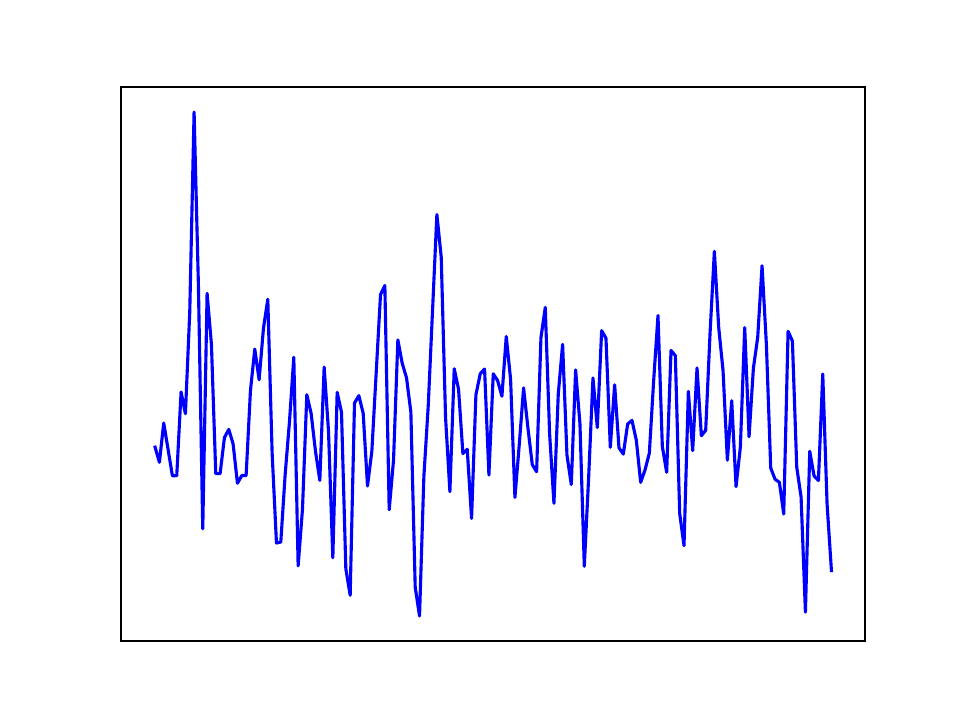}
\label{fig:cir_syn_a}}
\subfloat[Synthesized NLoS signal]{\includegraphics[height=1.3in]{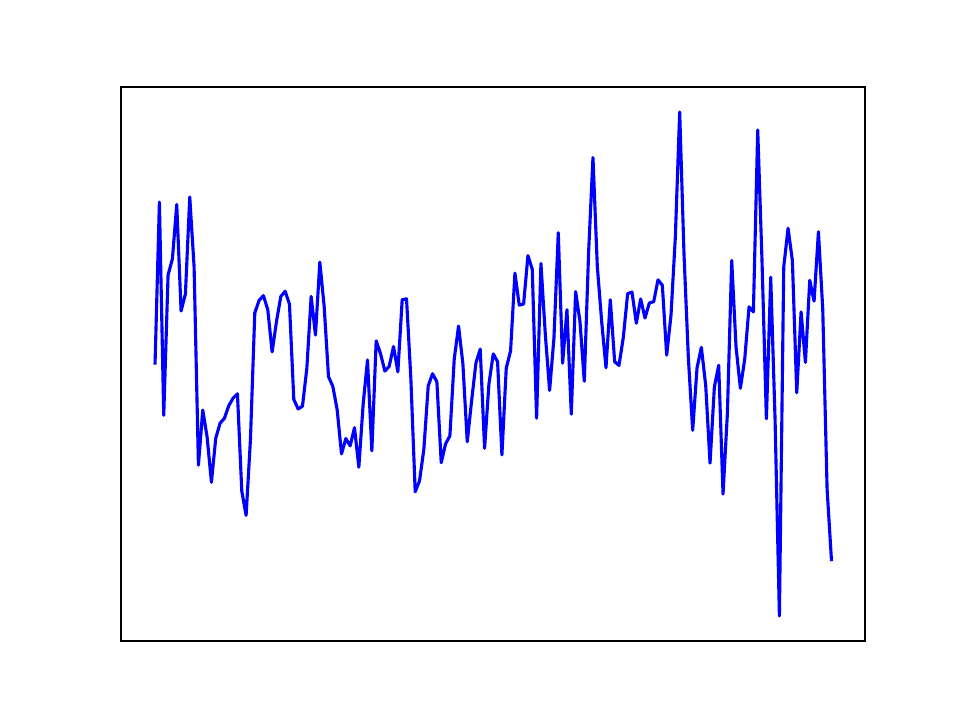}
\label{fig:cir_syn_b}}
\subfloat[Synthesized indoor signal]{\includegraphics[height=1.3in]{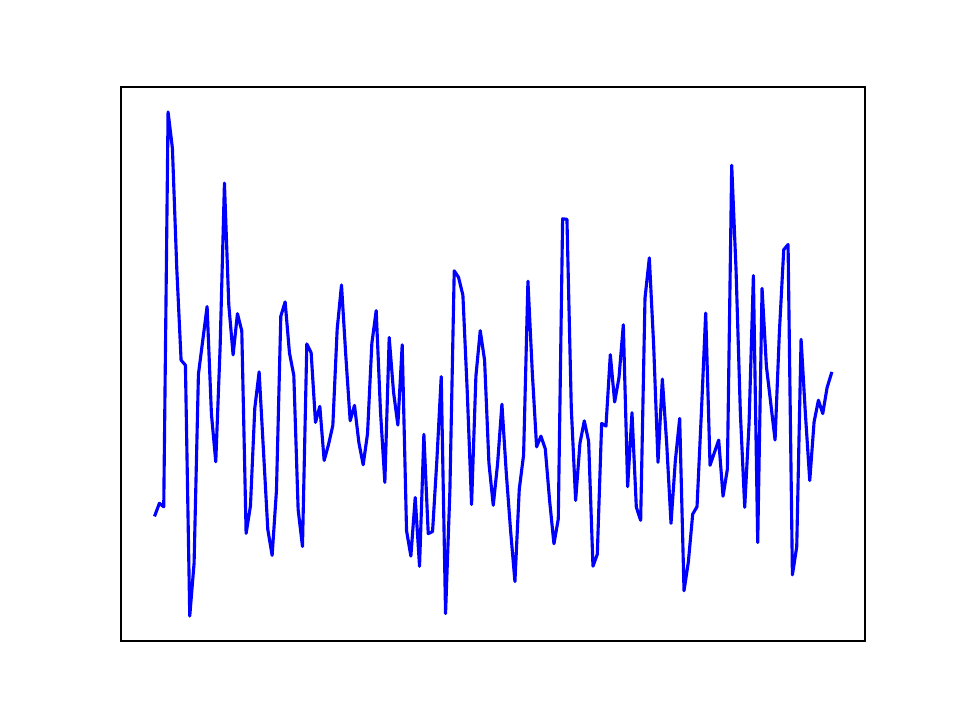}
\label{fig:cir_syn_c}}
\subfloat[Synthesized outdoor signal]{\includegraphics[height=1.3in]{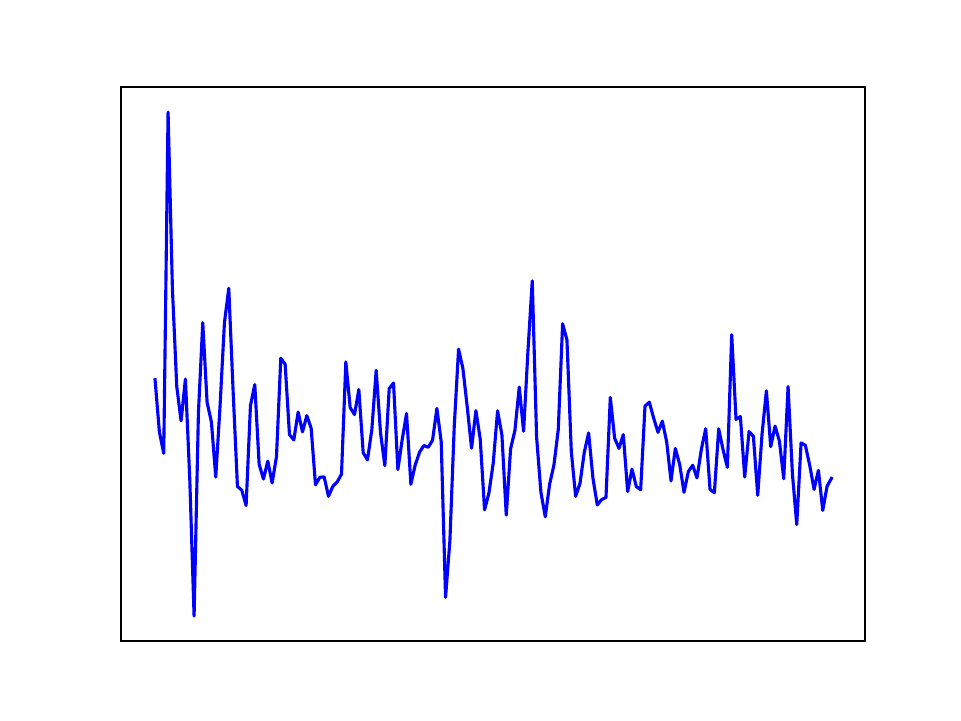}
\label{fig:cir_syn_d}} \\
\subfloat[Translated LoS signal]{\includegraphics[height=1.3in]{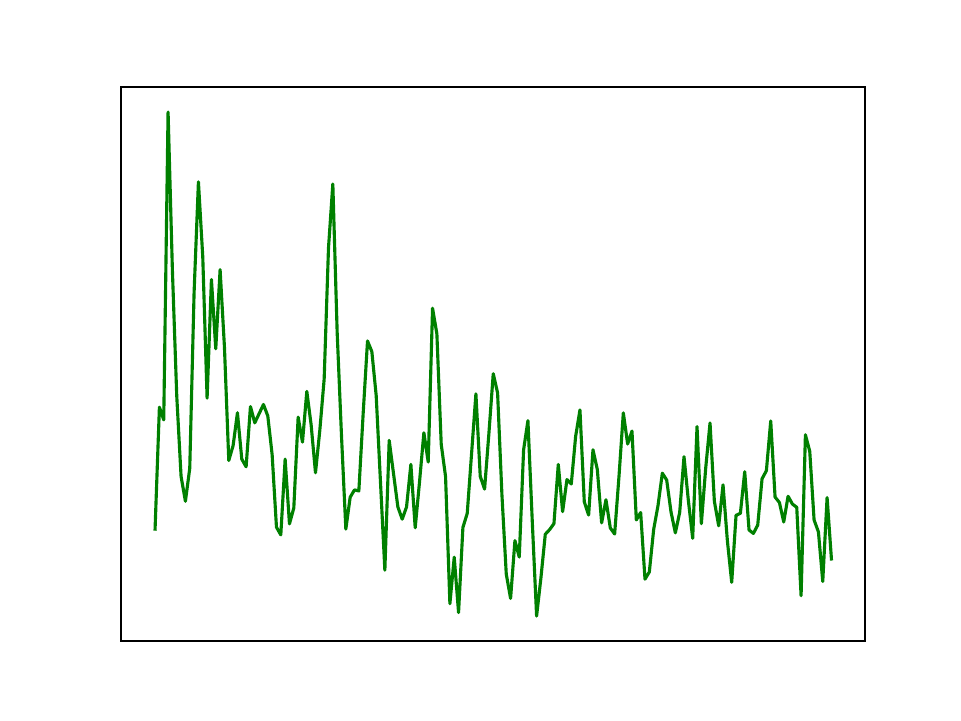}
\label{fig:cir_trans_a}}
\subfloat[Translated NLoS signal]{\includegraphics[height=1.3in]{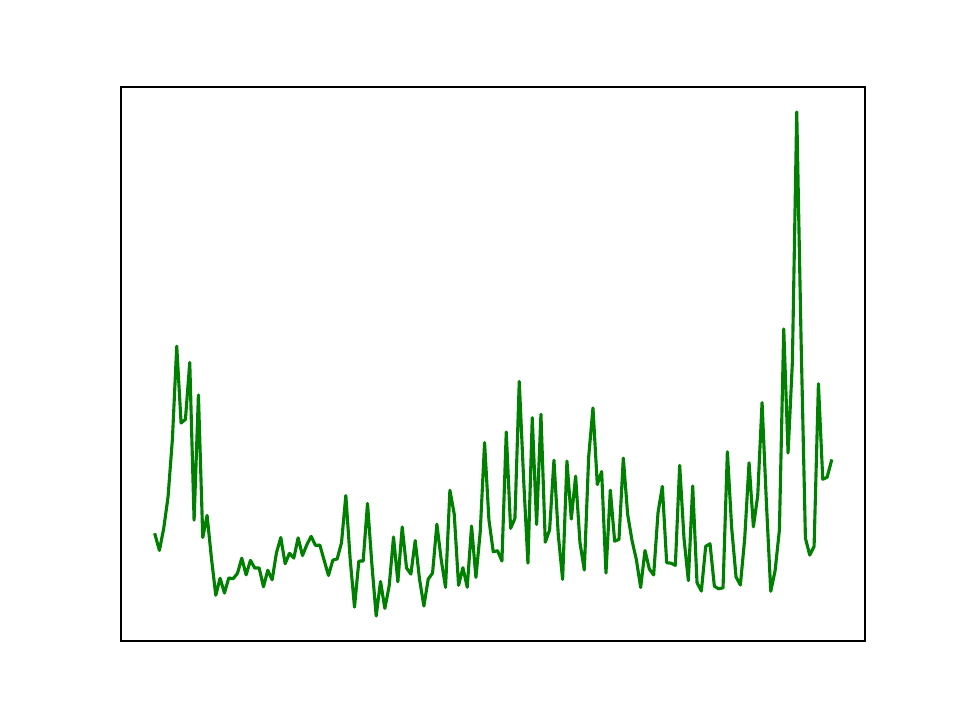}
\label{fig:cir_trans_b}}
\subfloat[Translated indoor signal]{\includegraphics[height=1.3in]{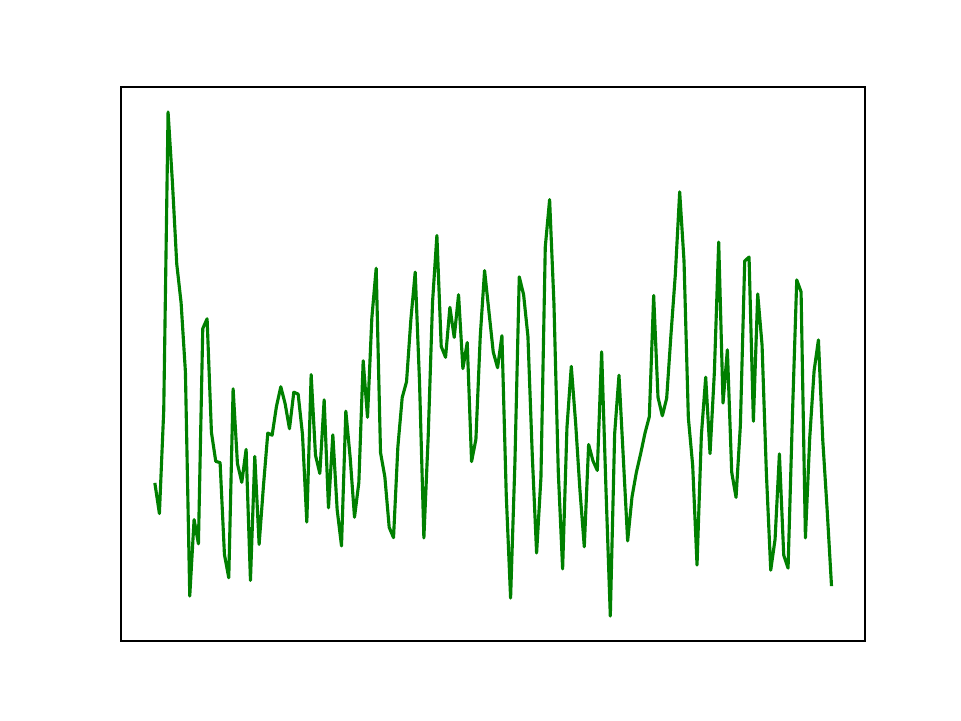}
\label{fig:cir_trans_c}}
\subfloat[Translated outdoor signal]{\includegraphics[height=1.3in]{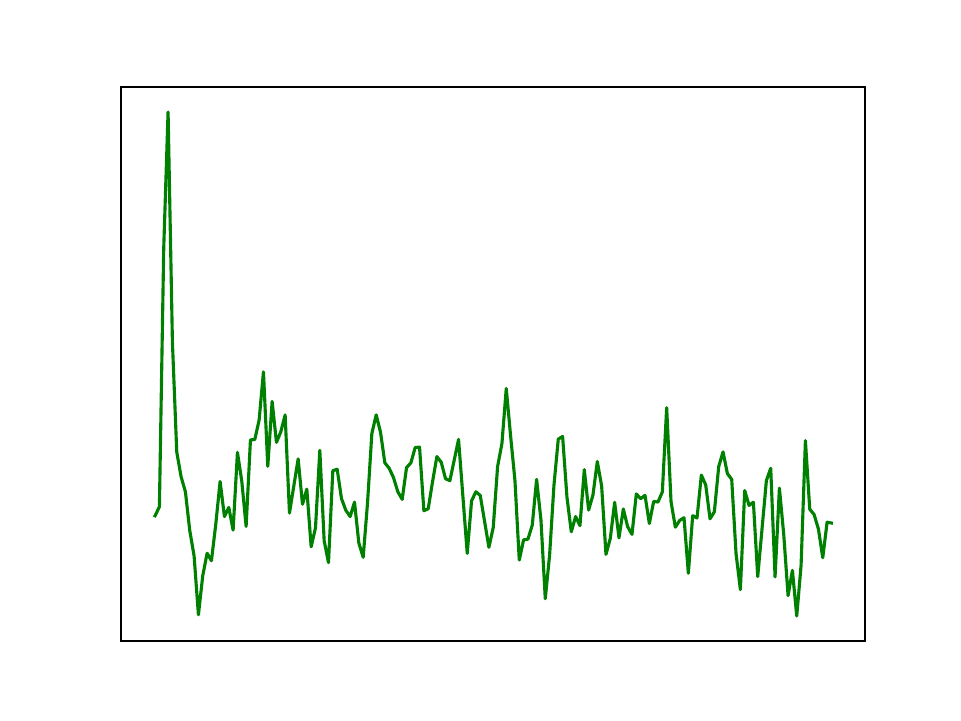}
\label{fig:cir_trans_d}}
\caption{The Signal Waveform of real, synthesized and translated signals.}
        \label{fig:CIRs}
\end{figure*}  

\begin{figure}[!t]
\centering
\subfloat[PSD between real and synthesized signals.]{\includegraphics[height=1.5in]{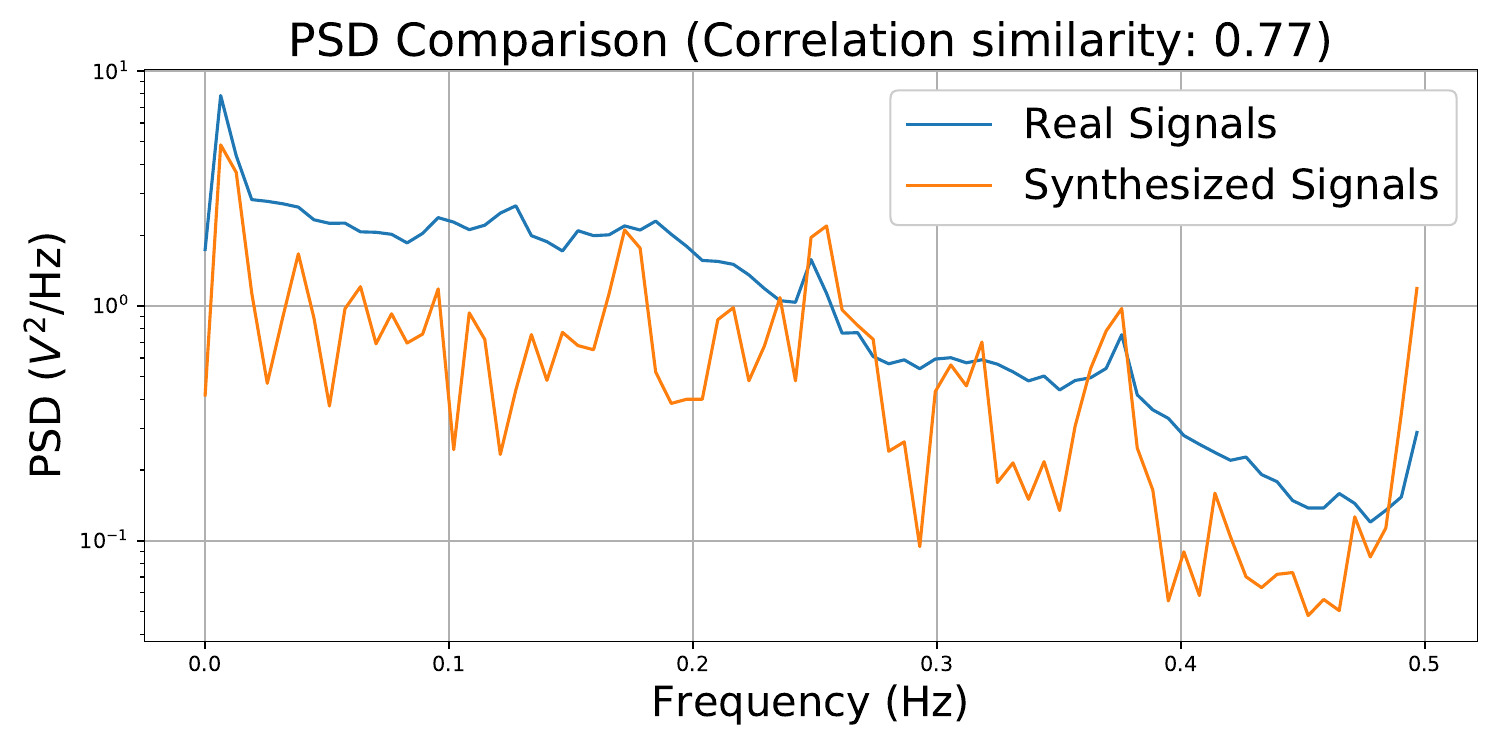}
\label{fig:psd_a}} \\
\subfloat[PSD between real and translated signals.]{\includegraphics[height=1.5in]{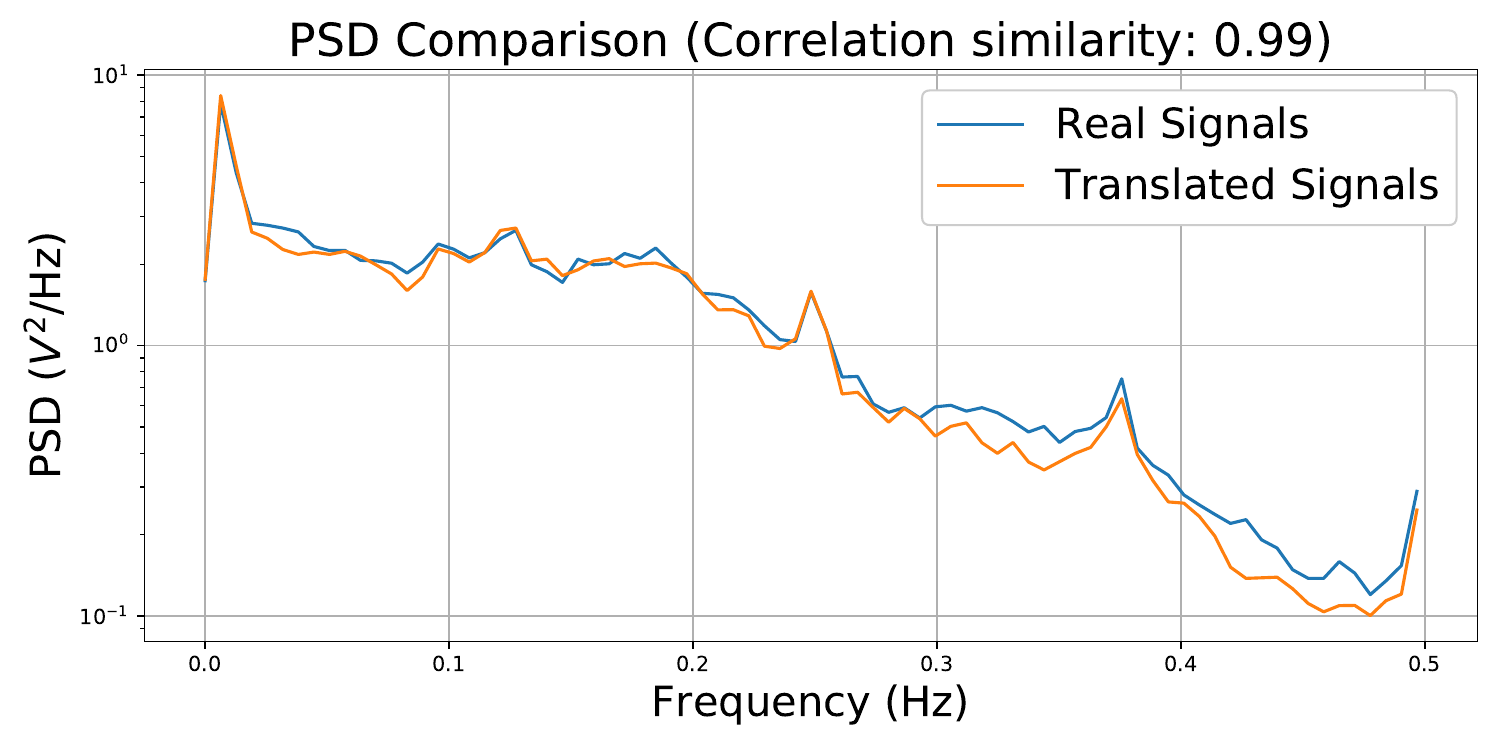}
\label{fig:psd_b}}
\caption{The PSD comparisons between real and two types of generated signals.}
\label{fig:PSD}
\end{figure}  

We first present the comprehensive comparisons between the real and two types of generated signal waveforms.
In Fig.~\ref{fig:CIRs}, we present a visualization of the real, synthesized, and translated signal waveforms under different environment scenarios.
The results demonstrate distinct characteristics of signals under different conditions. Specifically, the first path (FP) of the LOS signals exhibited a cleaner and more direct waveform, indicative of a straightforward signal path with minimal obstructions. In contrast, NLOS signals were characterized by a more complex FP, often distorted due to the presence of obstacles, thereby affecting the signal's integrity. Furthermore, the multipath components of the signals revealed additional disparities between indoor and outdoor settings. Indoor environments, laden with walls and objects, resulted in a higher density of multipath components (MPCs), leading to a more intricate signal waveform. On the other hand, outdoor signals displayed fewer MPCs, owing to the open space and reduced signal reflections. 

Next, we compare the average power spectrum density (PSD) of the real signals and the generated signals. Specifically, the real and label-based synthesized signal PSDs are in Fig.~\ref{fig:psd_a}, while the signal-based translated signal PSDs are in Fig.~\ref{fig:psd_b}.
The average correlation similarity (measured in cosine similarity) of the two signals is also marked in each figure. Specifically, the correlation between real and synthesized signals is 0.77, while the correlation between real and translated signals is 0.99. The results show that both types of the generated signals are highly similar to the real signals on average, while the translated signals achieve a better performance.

These visualizations provide a qualitative assessment of the ability of our framework to generate realistic signals.
The visualization results demonstrate that the proposed techniques can generate signals closely matching real-world measurements. This highlights the effectiveness of our framework in realistic wireless signal generation.

\subsection{Physical Feature Comparison}
\label{sec:e_physical_feature}

\begin{table*}[!t]
\caption{Comparison of real and synthesized signals on four physical features}
\label{tab:PF_syn}
\begin{center}
\begin{small}
\begin{sc}
\begin{tabular}{lcccc}
\toprule
{Data Type} & {Maximum Amplitude (MA)} & 
 {Rise Time (RT)} & {Energy} & {Kurtosis} \\
\midrule
{Synthesized General} & {0.87} & {0.44} & {-0.45} & {-0.09}  \\
\midrule
{Synthesized $d_1$} & {0.88} & {0.42} & {-0.46} & {-0.10} \\
{Synthesized $d_2$} & {0.87} & {0.46} & {-0.44} & {-0.07} \\
\midrule
{Synthesized LoS} & {0.81} & {0.43} & {-0.41} & {-0.01} \\
{Synthesized NLoS} & {0.90} & {0.44} & {-0.48} & {0.03} \\
{Synthesized indoor} & {0.83} & {0.65} & {-0.42} & {0.01} \\
{Synthesized outdoor} & {0.90} & {0.44} & {-0.45} & {-0.10} \\
\bottomrule
\end{tabular}
\end{sc}
\end{small}
\end{center}
\end{table*}

\begin{table*}[!t]
\caption{Comparison of real and translated signals on four physical features}
\label{tab:PF_trans}
\begin{center}
\begin{small}
\begin{sc}
\begin{tabular}{lcccc}
\toprule
{Data Type} & {Maximum Amplitude (MA)} & 
 {Rise Time (RT)} & {Energy} & {Kurtosis} \\
\midrule
{Translated General} & {0.96} & {0.59} & {0.87} & {0.14}  \\
\midrule
{Translated $d_1$} & {0.96} & {0.54} & {0.86} & {0.14} \\
{Translated $d_2$} & {0.97} & {0.63} & {0.87} & {0.15} \\
\midrule
{Translated LoS} & {0.97} & {0.92} & {0.86} & {0.06} \\
{Translated NLoS} & {0.97} & {0.50} & {0.87} & {0.08} \\
{Translated indoor} & {0.99} & {0.81} & {0.91} & {-0.29} \\
{Translated outdoor} & {0.96} & {0.52} & {0.87} & {0.19} \\
\bottomrule
\end{tabular}
\end{sc}
\end{small}
\end{center}
\end{table*}



Wireless signals possess unique characteristics that make them challenging to differentiate through human perception, yet these characteristics are crucial for wireless applications. To capture the difference between the real and generated signals more effectively, we extract PFs from these signals as suggested in \cite{WymMarGifWin:J12}. We conduct comparisons of four physical features including maximum amplitude (MA), rise time (RT), energy, and kurtosis in terms of cosine similarity, presented in Table~\ref{tab:PF_syn}. 

The histograms reveal distinct PF characteristics for LoS and NLoS signals, irrespective of whether they are real or generated. This observation underscores that the generated signals possess features akin to those of real signals, further attesting to the effectiveness of our proposed framework in generating realistic wireless signals.

\subsection{Data Augmentation Evaluation on Toy Models}
\label{sec:e_dist}

We evaluate the utility of the synthesized and translated signals for training toy machine learning models for distance estimation and environment identification tasks. The results demonstrate the potential of our framework for data augmentation. We employ convolutional neural network (CNNs) as our toy models. 

The toy models are trained on datasets composed of real, synthesized, or translated signals. In particular, we construct two sets of synthesized signals, one of the same size of the real signals while one of 10 times the size of the real signals, denoted as "synthesized (x1)" and "synthesized (x10)" in Table~\ref{tab:toy_classifier}. The same for the translated signals. The real data is from the toy-model training set aside from the general training set. After training, the toy models are evaluated on the testing set, as introduced in section \ref{sec:e_data}.

\begin{table}[!t]
\caption{Performances of toy CNN on distance estimation, trained on real, synthesized and translated signals.}
\label{tab:toy_estimator}
\begin{center}
\begin{small}
\begin{sc}
\begin{tabular}{lcc}
\toprule
{Data Type} & {MSE (m)} & 
 {MAE (m)} \\
\midrule
{Real} & {0.17} & {0.11}  \\
\midrule
{Synthesized ($\times 1$)} & {0.19} & {0.12} \\
{Synthesized ($\times 10$)} & {0.14} & {0.09} \\
\midrule
{Translated ($\times 1$)} & {0.17} & {0.11} \\
{Translated ($\times 10$)} & {0.14} & {0.09} \\
\bottomrule
\end{tabular}
\end{sc}
\end{small}
\end{center}
\end{table}

\begin{table}[!t]
\caption{Performances of toy CNN on environment identification, trained on real, synthesized and translated signals.}
\label{tab:toy_classifier}
\begin{center}
\begin{small}
\begin{sc}
\begin{tabular}{lcc}
\toprule
\multirow{2}{*}{Data Type} & \multirow{2}{*}{\shortstack{LoS/NLoS \\ Acc}} & 
 \multirow{2}{*}{\shortstack{Indoor/Outdoor \\ Acc}} \\
 & & \\
\midrule
{Real} & {0.74} & {0.67}  \\
\midrule
{Synthesized ($\times 1$)} & {0.74} & {0.64} \\
{Synthesized ($\times 10$)} & {0.75} & {0.74} \\
\midrule
{Translated ($\times 1$)} & {0.74} & {0.71} \\
{Translated ($\times 10$)} & {0.75} & {0.75} \\
\bottomrule
\end{tabular}
\end{sc}
\end{small}
\end{center}
\end{table}

\subsubsection{Training CNNs for Distance Estimation}

First, we train separate CNNs on the real and four sets of generated signals for distance estimation. These CNNs, referred to as toy CNN regressors, take in signal measurements and produce distance estimates.
We utilize the root mean square error (RMSE) and the mean absolute error (MAE) to evaluate the estimation performance. The results are shown in Table~\ref{tab:toy_estimator}.

\subsubsection{Training CNNs for environment Identification}

\begin{table}[!t]
\caption{Performances before translation.}
\label{tab:toy_classifier}
\begin{center}
\begin{small}
\begin{sc}
\begin{tabular}{lcc}
\toprule
\multirow{2}{*}{Data Type} & \shortstack{LoS} & 
 \shortstack{NLoS} \\
 & & \\
\midrule
{LoS} & {0.9} & {0.7}  \\
\midrule
{NLoS} & {0.7} & {0.9} \\
\bottomrule
\end{tabular}
\end{sc}
\end{small}
\end{center}
\end{table}

\begin{table}[!t]
\caption{Performances after translation.}
\label{tab:toy_classifier}
\begin{center}
\begin{small}
\begin{sc}
\begin{tabular}{lcc}
\toprule
\multirow{2}{*}{Data Type} & \shortstack{LoS} & 
 \shortstack{NLoS} \\
 & & \\
\midrule
{LoS (translated)} & {0.7} & {0.9}  \\
\midrule
{NLoS (translated)} & {0.9} & {0.7} \\
\bottomrule
\end{tabular}
\end{sc}
\end{small}
\end{center}
\end{table}

Next, we train separate CNNs on the real and four sets of generated signals for environment identification. These CNNs, referred to as toy classifiers, ingest signal measurements and output environment labels. The results are shown in Table~\ref{tab:toy_classifier}.
The results indicate that the use of generated data for data augmentation enhances the performance of learning-based methods. This underscores the value of our proposed framework in generating realistic wireless signals for data augmentation, thereby improving the effectiveness of machine learning models in wireless sensing tasks.




\section{Conclusion}
\label{sec:con}

In this paper, we introduced IIns-GAN, a DL-based method to generate realistic wireless signal with position-related labels. This method interpreted labeled signal generation as an inference problem on a LVM. The generation with target labels are then conducted by VI techniques and implemented through a GAN-based network.
Experimental results show the promise of generated signals in a wide range of applications, including performance evaluation for wireless systems and data augmentation for learning-based algorithms. This hence provides a potential solution to challenges of realistic dataset acquisition in wirless sensing tasks.
Interpreted as a variational inference problem, the proposed method also provides a promising direction for the development of hybrid inference and DL techniques for complex signal processing problems.
Our future work will promote the scalability of the approach towards signal generation with more delicate environmental features.
This will enable the adaptation of our approach to various wireless systems and environments, making it more versatile and applicable in practical scenarios.

\ifCLASSOPTIONcaptionsoff
  \newpage
\fi

\end{document}